\documentclass[preprint]{elsarticle}
\def\FRESH{\textbf{F}eatu\textbf{R}e \textbf{E}xtraction based on \textbf{S}calable \textbf{H}ypothesis tests}

\def\citePoster{\cite{christ_time_2016}} 
\def\iPRODICT{iPRODICT}
\def\iPRODICTproduct{steel billets}
\def\iPRODICTgrant{01IS14004}
\newcommand{\alg}[1]{\texttt{#1}}

\newcommand{\ftr}[1]{\texttt{#1}($S$)}
\newcommand{\mathftr}[1]{\texttt{#1}(S)}
\newcommand{\ftrpar}[2]{\texttt{#1}($S|#2$)}

\usepackage{picins}

\usepackage{graphicx}
\graphicspath{{./graphics/}}
\usepackage{algorithm2e}
\usepackage{array}
\journal{Neurocomputing}

\usepackage[utf8]{inputenc}				
\usepackage[T1]{fontenc}					
\usepackage[english]{babel}
\addto\captionsenglish{\renewcommand{\appendixname}{Appendix }}

\usepackage{amsmath}

\DeclareMathOperator{\median}{median}
\DeclareMathOperator{\mean}{mean}
\DeclareMathOperator{\var}{var}
\usepackage{amssymb}					
\usepackage{subfigure}
\usepackage{hyperref}

\makeatletter
\providecommand{\doi}[1]{%
  \begingroup
    \let\bibinfo\@secondoftwo
    \urlstyle{rm}%
    \href{http://dx.doi.org/#1}{%
      doi:\discretionary{}{}{}%
      \nolinkurl{#1}%
    }%
  \endgroup
}
\makeatother

\usepackage{enumitem}					

\newtheorem{defi}{Definition}

\usepackage{multirow}

\usepackage{rotating}
\usepackage{longtable}
\usepackage[version-1-compatibility]{siunitx} 

\usepackage[table]{xcolor}

\begin{document}

\begin{frontmatter}

\title{Distributed and parallel time series feature extraction for industrial big data applications\tnoteref{t1}}
\tnotetext[t1]{The feature extraction algorithms and the \alg{FRESH} algorithm itself, all of which are described in this work, have been implemented in a open source python package called \alg{tsfresh}. Its source code be found at \url{https://github.com/blue-yonder/tsfresh}.}

\author[a]{Maximilian Christ}
\ead{maximilian.christ@blue-yonder.com}

\author[b,c]{Andreas W. Kempa-Liehr}
\ead{a.kempa-liehr@auckland.ac.nz}
\cortext[*]{Corresponding author}

\author[a,d]{Michael Feindt}
\ead{michael.feindt@blue-yonder.com}
\fntext[d]{on leave of absence from Karlsruhe Institute of Technology}

\address[a]{Blue Yonder GmbH, Karlsruhe, Germany}
\address[b]{Department of Engineering Science, University of Auckland, Auckland, New Zealand}
\address[c]{Freiburg Materials Research Center, University of Freiburg, Freiburg, Germany}

\begin{abstract}
The all-relevant problem of feature selection is the identification of all strongly and weakly relevant attributes.
This problem is especially hard to solve for time series classification and regression in industrial applications such as predictive maintenance or production line optimization, for which each label or regression target is associated with several time series and meta-information simultaneously.
Here, we are proposing an efficient, scalable feature extraction algorithm for time series, which filters the available features in an early stage of the machine learning pipeline with respect to their significance for the classification or regression task, while controlling the expected percentage of selected but irrelevant features.
The proposed algorithm combines established feature extraction methods with a feature importance filter.
It has a low computational complexity, allows to start on a problem with only limited domain knowledge available, can be trivially parallelized, is highly scalable and based on well studied non-parametric hypothesis tests.
We benchmark our proposed algorithm on all binary classification problems of the UCR time series classification archive as well as time series from a production line optimization project and simulated stochastic processes with underlying qualitative change of dynamics.

\end{abstract}


\end{frontmatter}

\section{Introduction}
\label{section_introduction}

Utilizing the continuously increasing amount of accessible data provides tremendous benefit for the understanding of the complex systems, which we are part of and contribute to.
Examples range from the often counter-intuitive behavior of social systems and their macro-level collective dynamics \cite{Helbing2015_complexity} to disease dynamics \cite{Wang2016_vaccination} and precision medicine \cite{CollinsVarmus2015_DS}.
Other promising fields of application for machine learning are the Internet of Things (IoT) \citep{gubbi_internet_2013}
and Industry 4.0 \citep{hermann_design_2016} environments.
In these fields, machine learning models anticipate future device states by combining knowledge about device attributes with historic sensor time series.
They permit the classification of devices (e.g. hard drives) into risk classes with respect to a specific defect \citep{mobley_introduction_2002}. 
Both fields are driven by the availability of cheap sensors 
and advancing connectivity between devices, which increases the need for machine learning on temporally annotated data.

In most cases the volume of the generated time series data forbids their transport to centralized databases \citep{gubbi_internet_2013}.
Instead, algorithms for an efficient reduction of data volume by means of feature extraction and feature selection are needed \cite[p. 125--136]{bolon-canedo_feature_2015}.
Furthermore, for online applications of machine learning it is important to continuously select relevant features in order to deal with concept drifts caused by qualitative changes of the underlying dynamics \citep{LiuSetiono1998_fs}.

Therefore, for industrial and other applications, one needs to combine distributed feature extraction methods with a scalable feature selection, especially for problems where several time series and meta-information have to be considered per label/target \citep{kusiak_prediction_2011}.
For time series classification, it proved to be efficient to apply comprehensive feature extraction algorithms and then filter the respective features \citep{fulcher_highly_2014}.

Motivated by industrial applications for machine learning models \citePoster{}, we are extending the approach of \citeauthor{fulcher_highly_2014} \citep{fulcher_highly_2014} by a highly parallel feature filtering and propose \FRESH{} (\alg{FRESH}). 
The algorithm characterizes time series with comprehensive and well-established feature mappings and considers additional features describing meta-information. 
In a second step, each feature vector is individually and independently evaluated with respect to its significance for predicting the target under investigation. 
The result of these tests is a vector of p-values, quantifying the significance of each feature for predicting the label/target. 
This vector is evaluated on basis of the Benjamini-Yekutieli procedure \citep{benjamini_control_2001} in order to decide which features to keep. 

The proposed algorithm is evaluated on all binary classification problems of the UCR time series classification archive \citep{chen_ucr_2015} as well as time series data from a production line optimization project and simulated time series from a stochastic process with underlying qualitative change of dynamics \citep[p. 164]{liehr_dissipative_2013}.
The results are benchmarked against well-established feature selection algorithms like linear discriminant analysis \citep{fulcher_highly_2014} and the Boruta algorithm \citep{kursa_all_2011}, but also against Dynamic Time Warping \citep{wang_experimental_2013}.
The analysis shows that the proposed method outperforms Boruta based feature selection approaches as well as Dynamic Time Warping based approaches for problems with large feature sets and large time series samples.

The \alg{FRESH} algorithm and the comprehensive framework to extract the features have also been implemented in a Python package called \alg{tsfresh}.
Its source is open, and it can be downloaded from \url{www.github.com/blue-yonder/tsfresh}.
During the writing of this paper, \alg{tsfresh} is the most comprehensive feature extraction package that is publicly available in Python.
The package is designed for the fast extraction of a huge number of features while being compatible with popular Python machine learning frameworks such as scikit-learn \cite{pedregosa_scikitlearn_2011}, numpy \cite{van2011numpy} or pandas \cite{mckinney-proc-scipy-2010}.

This work starts with an introduction to time series feature extraction in Sec.~\ref{sec:subsec_mapping}. 
Afterwards, in Sec.~\ref{sec:feature_filtering}, the \alg{FRESH} algorithm is introduced. 
In Sec.~\ref{Sec:evaluation}, the performance of its Python implementation is evaluated on the UCR time series and a industrial dataset.
Afterwards, characteristics of \alg{FRESH} are discussed in Sec.~\ref{sec:discussion}.
This work closes with a summary and an outlook over future work in in Sec.~\ref{sec:summary}.
Additionally, \ref{list_of_feature_mappings} contains an overview over the considered feature mappings.

\section{Time series feature extraction}
\label{sec:subsec_mapping}

\subsection{Time series}
Temporally annotated data come in three different variants \citep{ElmasriLee1998_temporal}: 
Temporally invariant information (e.g. the manufacturer of a device), temporally variant information, which change irregularly (e.g. process states of a device), and temporally variant information with regularly updated values (e.g. measurements of sensors on a device). 
The latter describe the continuously changing state $s_{i,j}(t)$ of a system or device $s_i$ with respect to a specific measurement of sensor $j$, which is repeated in intervals of length $\Delta_t$. 
This sampling captures the state of the system or device under investigation as a sequence of length $n_t^{(j)}$
\begin{subequations}\label{ts_sampling}
\begin{equation}
s_{i,j}(t_1) \rightarrow s_{i,j}(t_2) \rightarrow \ldots \rightarrow s_{i,j}(t_\nu) \rightarrow \ldots \rightarrow s_{i,j}(t_{n_t^{(j)}})
\end{equation}
with $t_{\nu+1} = t_\nu + \Delta_t$.
Such kind of sequences are called time series\footnote{The \alg{FRESH} algorithm was developed while having time series in mind where $t_\nu$ denotes a point in time.
However, it can also be applied on other uniformly sampled signals such as spectra, e.g. signals where $t_\nu$ denotes a wave length.} 
and are abbreviated by
\begin{equation}
\begin{split}
S_{i,j}&=(s_{i,j}(t_1), s_{i,j}(t_2), \ldots, s_{i,j}(t_\nu),\ldots,s_{i,j}(t_{n_t^{(j)}}))^{\text{T}}\\
&=  (s_{i,j,1}, s_{i,j,2}, \ldots, s_{i,j,\nu},\ldots,s_{i,j,n_t^{(j)}})^{\text{T}}.
\end{split}
\end{equation}
\end{subequations}
Here, we are considering $i=1,\ldots,m$ devices with recordings for $j=1,\ldots, n$ different time series.
The recorded time series of one type $j$ should have the same same length $n_t^{(j)}$ for all devices.
So in total we are dealing with $n\cdot m\cdot \sum_{j=1}^m n_t^{(j)}$ values describing the input for the multi-variate time series problem under investigation. 

\subsection{Time series classification}

We assume that we are given a target vector $Y = (y_1, \ldots, y_m)^T$ where the entry $y_i$ of $Y$ describes a characteristic or state of device $i$.
Those can either be represented by discrete or continuous values.
In the first case, the prediction of the values from $Y$ based on the different time series $S = (S_{i,j})_{i=1,\ldots,m, j=1,\ldots, n}$ states a multi-variate time series classification problem, in the later we perform a regression on multi-variate time series input.

The first problem class, which deals with the assigning of time series to discrete categories is called \emph{time series classification} (TSC).
There are different approaches to tackle TSC problems; \citet{bagnall2016great} give an overview over common approaches and recent algorithmic advances in the field of TSC.

Shape-based approaches identify similar pairs of time series in terms of their values through time.
For this purpose a similarity metrics like the Euclidean distance is applied and algorithms like $k$-nearest-neighbors (kNN) can be used to find similar time series and their associated target values $\{y_1,...,y_k\}$.
Then, the class label is for example predicted as the majority vote of across the $k$ neighbors \cite{wang_characteristicbased_2006}.
Dynamic Time Warping (DTW) is a variant of shape-based approaches, which accommodates local (temporal) shifts in the data \cite{muller_dynamic_2007}), 

On the other hand, direct approaches learn a representation for the input objects.
Neural networks and their deep extensions deploy a network of neurons and weights, being especially suited to applications where manual feature engineering is difficult.
Common extensions like recurrent neural networks (RNN) \cite{husken2003recurrent}, stacked restricted Boltzmann-machines (RBM) \cite{lee2009unsupervised}, or convolutional neural networks (CNN) \cite{yang_deep_2015} have already been deployed for TSC tasks.

A third group of TSC approaches is feature-based, which will be discussed in the following section.

\subsection{Feature mapping}

In this work, in order to characterize a time series and reduce the data volume, a mapping $\theta_k:\mathbb{R}^{n_t^{(j)}}\rightarrow\mathbb{R}$ is introduced, which captures a specific aspect of the time series. 
One example for such mapping might be the maximum operator
\begin{equation*}
\theta_{\max}(S_{i,j}) = \max \{s_{i,j,1}, s_{i,j,2}, \ldots, s_{i,j,\nu},\ldots,s_{i,j,n_t^{(j)}}\},
\end{equation*}
which quantifies the maximal value ever recorded for time series $S_{i,j}$.
This kind of lower dimensional representation is called a \emph{feature}, which is a measurable characteristics of the considered time series.
The feature mappings that we consider are stateless and cannot use information from other time series to derive its value  (this e.g. prohibits rankings of the time series with respect to a certain metric as a feature mapping).
Other examples for feature mappings $\theta_k$ of time series might be their mean, the number of peaks with a certain steepness, their periodicity, a global trend, etc. 

There is a comprehensive literature on the topic of time series feature extraction.
Early on, authors started to discuss the extraction of basic features such as max, min, skewness \cite{nanopoulos2001feature} or generic patterns such as peaks \cite{geurts2001pattern}.
Apart from that, every field that investigates time series, discusses specialized features for their applications, e.g. peak features to classify audio data \cite{mierswa_automatic_2005}, wavelet based features to monitor vibrations \cite{yen_wavelet_2000}, the parameters of a fitted exponential function to estimate the residual life of bearings \cite{gebraeel2004residual} or logarithmic periodogram features to detect arcs in the contact strips of high velocity trains \cite{barmada2014arc}. 
Finally, \citet{fulcher_highly_2014} and \citet{nun_fats_2015} build comprehensive collections of such time series feature mappings over different domains. 
\citet{fulcher_highly_2014} even collect more than 9000 features from 1000 different feature generating algorithms that are discussed in fields such as medicine, astrophysics, finance, mathematics, climate science, industrial applications and so on.
In this work, we will consider a smaller number of 111 features, a list of the considered mappings is given in \ref{list_of_feature_mappings}.

\begin{figure}[t]
 \includegraphics[width=\linewidth]{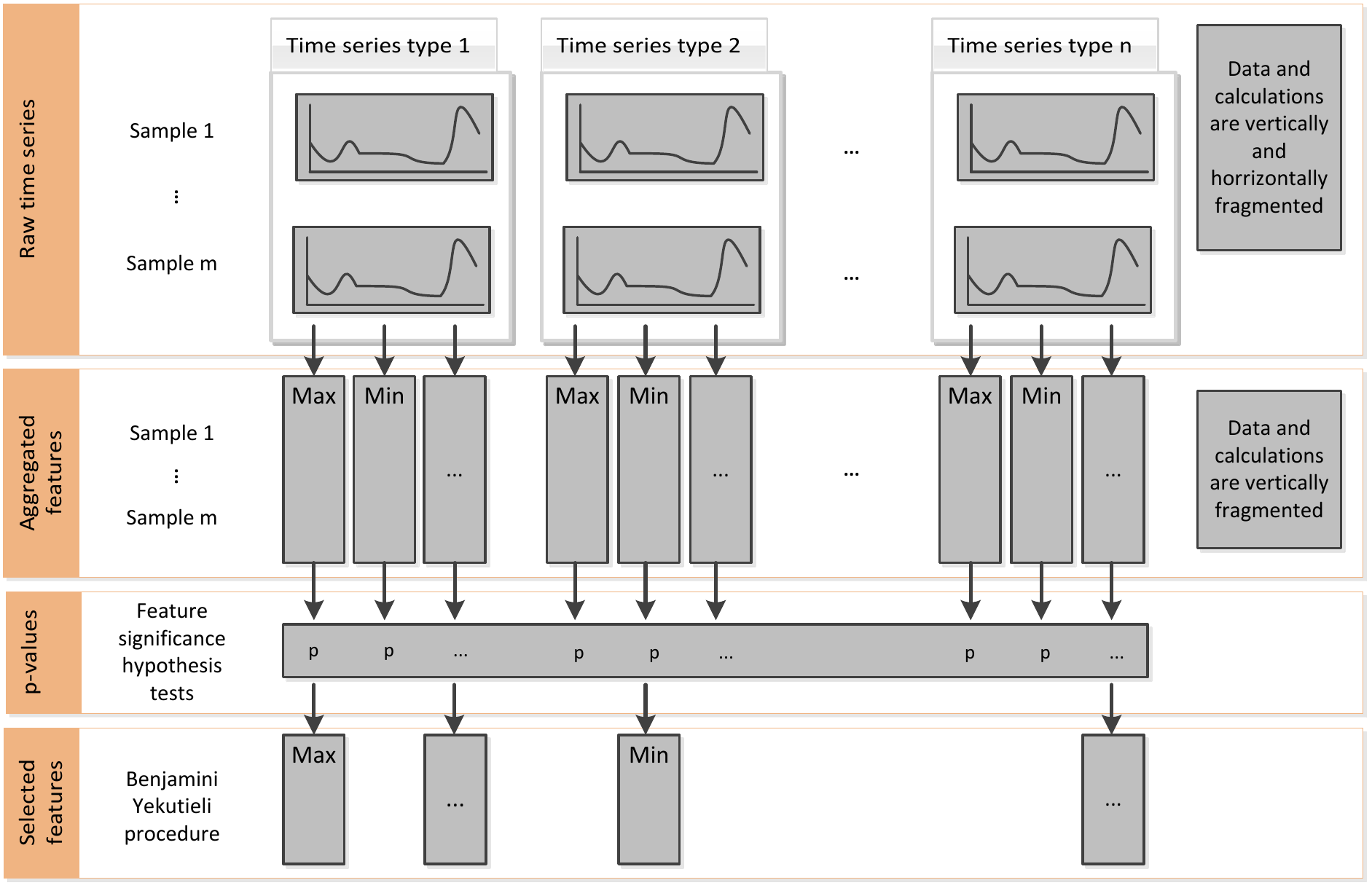}
 \caption{Data processing tiers of the filtered feature extraction algorithm.
 All but the Benjamini-Yekutieli procedure can be computed in parallel.}
  \label{fig:ffe_in_detail}
\end{figure}

Now, consider $n_f$ different time series feature mappings, which are applied to all $m\cdot n$ time series recorded from $n$ sensors of $m$ devices (Fig.~\ref{fig:ffe_in_detail}).
The resulting feature matrix $\mathbb{X}\in\mathbb{R}^{m\times n_\phi}$ has $m$ rows (one for each device) and $n_\phi = n \cdot n_f + n_i$ columns with $n_i$ denoting the number of features generated from device specific meta-information.
Each column of $\mathbb{X}$ comprises a vector $X\in\mathbb{R}^m$ capturing a specific characteristic of all considered devices.
The resulting feature matrix and the target vector are the base for supervised classification algorithms such as a Random Forest classifier.
By help of the feature matrix, the TSC task becomes a supervised classification problem.

Because it is such a common approach, many authors already tackled TSC tasks in a feature-based fashion
\cite{nanopoulos2001feature, geurts2001pattern, mierswa_automatic_2005, gebraeel2004residual, fulcher2014highly, xing2010brief}.
A very recent approach is the COTE algorithm \cite{bagnall2015time}, which computes 35 classifiers on four different data transformations capturing similarities in the time, frequency, change, and shape domains.
On the UCR archive, an important benchmark collection of TSC problems, it was able reach a higher accuracy than any other previously published TSC algorithm.

\section{Feature filtering}
\label{sec:feature_filtering}

Typically, time series are noisy and contain redundancies.
Therefore, one should keep the balance between extracting meaningful but probably fragile features and robust but probably non-significant features.
Some features such as the median value will not be heavily influenced by outliers, others such as the maximal value of the time series will be intrinsically fragile.
The choice of the right time series feature mappings is crucial to capture the right characteristics for the task at hand. 



\subsection{Relevance of features}


A meaningless feature describes a characteristic of the time series that is not useful for the classification or regression task at hand.
\citet{radivojac_feature_2004} measure the relevance of feature X for the classification of a binary target $Y$ as the difference between the class conditional distribution of $X$ given $Y=y_1$, denoted by $f_{X |Y = y_1}$, and of $X$ given $Y=y_2$, denoted by $f_{X |Y = y_2}$. 

We adopt this definition and consider a feature $X$ being relevant for the classification of binary target $Y$ if the two conditional density functions are not equal.
So, a feature $X$ will be denoted relevant for predicting target $Y$ if and only if
\begin{equation}
\label{eqn_relevant}
\exists \ y_1, y_2 \ \text{with} \ f_Y(y_1)>0,f_Y(y_2)>0 \ :\ f_{X |Y = y_1} \neq f_{X |Y = y_2}.
\end{equation}
Having different class conditional distribution as in Eq.~\eqref{eqn_relevant} is actually equivalent to $X$ and $Y$ being statistically dependent.
This becomes clear in the opposite case, when the feature $X$ is not relevant:
\begin{equation}
\begin{split}
\label{proof_equi_sig}
& \quad X \ \text{is not relevant for target} \ Y \\
\Leftrightarrow & \quad \forall y_1, y_2 \ \text{with} \ f_Y(y_1)>0, f_Y(y_2)>0: f_{X |Y = y_1} = f_{X |Y = y_2}  \\
\Leftrightarrow & \quad \forall y_1\ \text{with} \ f_Y(y_1)>0: f_{X | Y=y_1} =  f_{X} \\
\Leftrightarrow & \quad f_{X, Y} = f_{X | Y} f_Y = f_X f_Y \\
\Leftrightarrow  & \quad X \text{ and } Y \ \text{are statistically independent}\\
\end{split}
\end{equation}
We can therefore use the statistical independence to derive a shorter definition of a relevant feature:
\begin{defi}[A relevant feature]
A feature $X_\phi$ is \emph{relevant} or \emph{meaningful} for the prediction of $Y$ if and only if $X_\phi$ and $Y$ are not statistically independent.
\end{defi}
There are different approaches on how to check if a given feature is fulfilling this definition or not.
We will address the significance of a feature by means of hypothesis testing, a statistical inference technique \cite{rohatgi2015introduction}.

\subsection{Hypothesis tests}
\label{sec:subsec_featfilt}

\label{section_tests}

For every extracted feature $X_1,\ldots,X_\phi,\ldots,X_{n_\phi}$ we will deploy a singular statistical test checking the hypotheses 
\begin{equation}
\label{eqn_glb_hypo}
\begin{split}
H_0^\phi &= \{ X_\phi \text{ is irrelevant for predicting } Y \}, \\
H_1^\phi &= \{ X_\phi \text{ is relevant for predicting } Y \}.
\end{split}
\end{equation}
The result of each hypothesis test $H_0^\phi$ is a so-called p-value $p_\phi$, which quantifies the probability that feature $X_\phi$ is not relevant for predicting $Y$. 
Small p-values indicate features, which are relevant for predicting the target.

Based on the vector $(p_1,\ldots,p_{n_\phi})^\mathrm{T}$ of all hypothesis tests, a multiple testing approach will select the relevant features (Sec.~\ref{sec:subsec_multitest}). 
We propose to treat every feature uniquely by a different statistical test, depending on whether the codomains of target and feature are binary or not.
The usage of one general feature test for all constellations is not recommended. 
Specialized hypothesis tests yield a higher statistical power due to more assumptions about the codomains that can be used during the construction of those tests.
The proposed feature significance tests are based on nonparametric hypothesis tests, which do not make any assumptions about the distribution of the variables, thus ensuring robustness of the procedure.





\textbf{Exact Fisher test of independence:}
This feature significance test can be used if both the target and the inspected feature are binary.
Fisher's exact test \citep{fisher_interpretation_1922} is based on the contingency table formed by $X_{\phi}$ and $Y$.
It inspects if both variables are statistically independent, which corresponds to the hypotheses from Eq$.$~\eqref{eqn_glb_hypo}.
Fisher's test belongs to the class of exact tests.
For such tests, the significance of the deviation from a null hypothesis (e.g., the p-value) can be calculated exactly, rather than relying on asymptotic results.



\textbf{Kolmogorov-Smirnov test (binary feature):}
This feature significance test assumes the feature to be binary and the target to be continuous.
In general, the Kolmogorov-Smirnov (KS) test is a non-parametric and stable goodness-of-fit test, which checks if two random variables $A$ and $B$ follow the same distribution \citep{massey_kolmogorovsmirnov_1951}:
\begin{equation*}
H_0 = \{ f_{A} = f_{B} \}, \ H_1 = \{ f_{A} \neq f_{B}\}.
\end{equation*}
By conditionally modeling the distribution function of target $Y$ on the two possible values $x_1, x_2$ of the feature $X_\phi$ we can use the KS test 
to check if the distribution of $Y$ differs given different values of $X_\phi$.
Setting $A=Y| X_\phi=x_1$ and $B=Y| X_\phi=x_2$ results in
\begin{equation}
\label{enq_ks_hypo}
H_0^\phi = \{ f_{Y| X_\phi=x_1} = f_{Y| X_\phi=x_2} \}, \
H_1^\phi = \{ f_{Y| X_\phi=x_1} \neq f_{Y| X_\phi=x_2} \}.
\end{equation}

The hypotheses from Eq$.$~\eqref{eqn_glb_hypo} and \eqref{enq_ks_hypo} are equivalent as demonstrated in the chain of Eqns. \eqref{proof_equi_sig}.
Hence, the KS test can address the feature relevance of $X_\phi$.



\textbf{Kolmogorov-Smirnov test (binary target):}
When the target is binary and the feature non-binary, we can deploy the Kolmogorov-Smirnov test again.
We have to switch roles of target and feature variable, resulting in the testing of the following hypothesis:
\begin{equation*}
H_0^\phi = \{ f_{X_\phi| Y=y_1} = f_{X_\phi| Y=y_2} \}, \
H_1^\phi = \{ f_{X_\phi| Y=y_1} \neq f_{X_\phi| Y=y_2} \}.
\end{equation*}
This time $y_1$ and $y_2$ are the two possible values of $Y$ and $f_{X_\phi| Y=y_j}$ is the conditional density function of $X_\phi$ given $Y$.
This hypothesis is also equivalent to the one in Eq$.$~\eqref{eqn_glb_hypo}.


\textbf{Kendal rank test:}
This filter can be deployed if neither target nor feature are binary.
Kendall's rank test \citep{kendall_new_1938} checks if two continuous variables may be regarded as statistically dependent, hence naturally fitting our hypotheses from Eq$.$~\eqref{eqn_glb_hypo}.
It is a non-parametric test based on Kendall's rank statistic $\tau$, measuring the strength of monotonic association between $X_\phi$ and $Y$.
The calculation of the rank statistic is more complex when ties are involved \citep{adler_modification_1957}, i.e. feature or target are categorical. 

\subsection{Feature significance testing}
\label{sec:subsec_multitest}


\begin{figure}[tbp]
   \subfigure[Ordered p-Values]{\label{fig:by_fdr_big}%
     \includegraphics[width=0.6\linewidth]{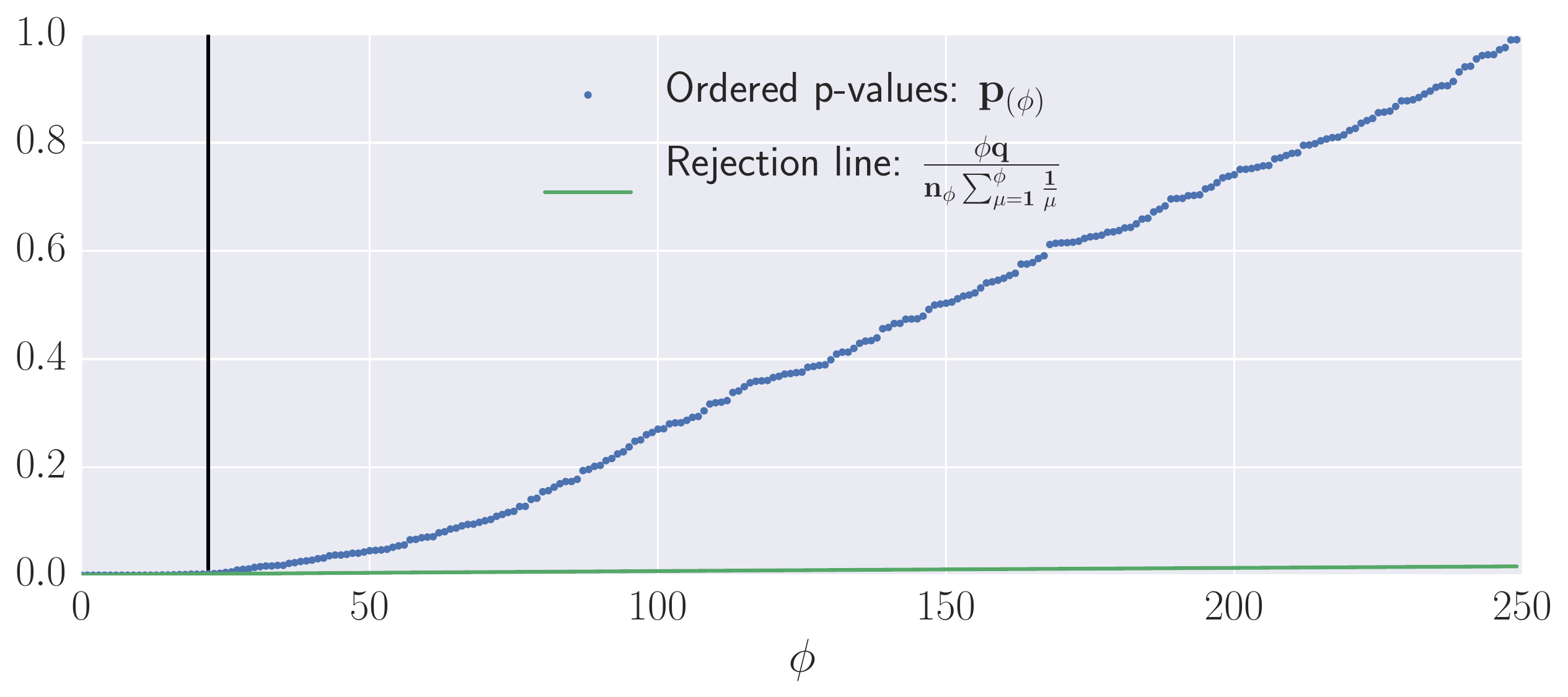}}%
   \
   \subfigure[The 40 lowest ordered p-values]{\label{fig:by_fdr_focus}%
     \includegraphics[width=0.37\linewidth]{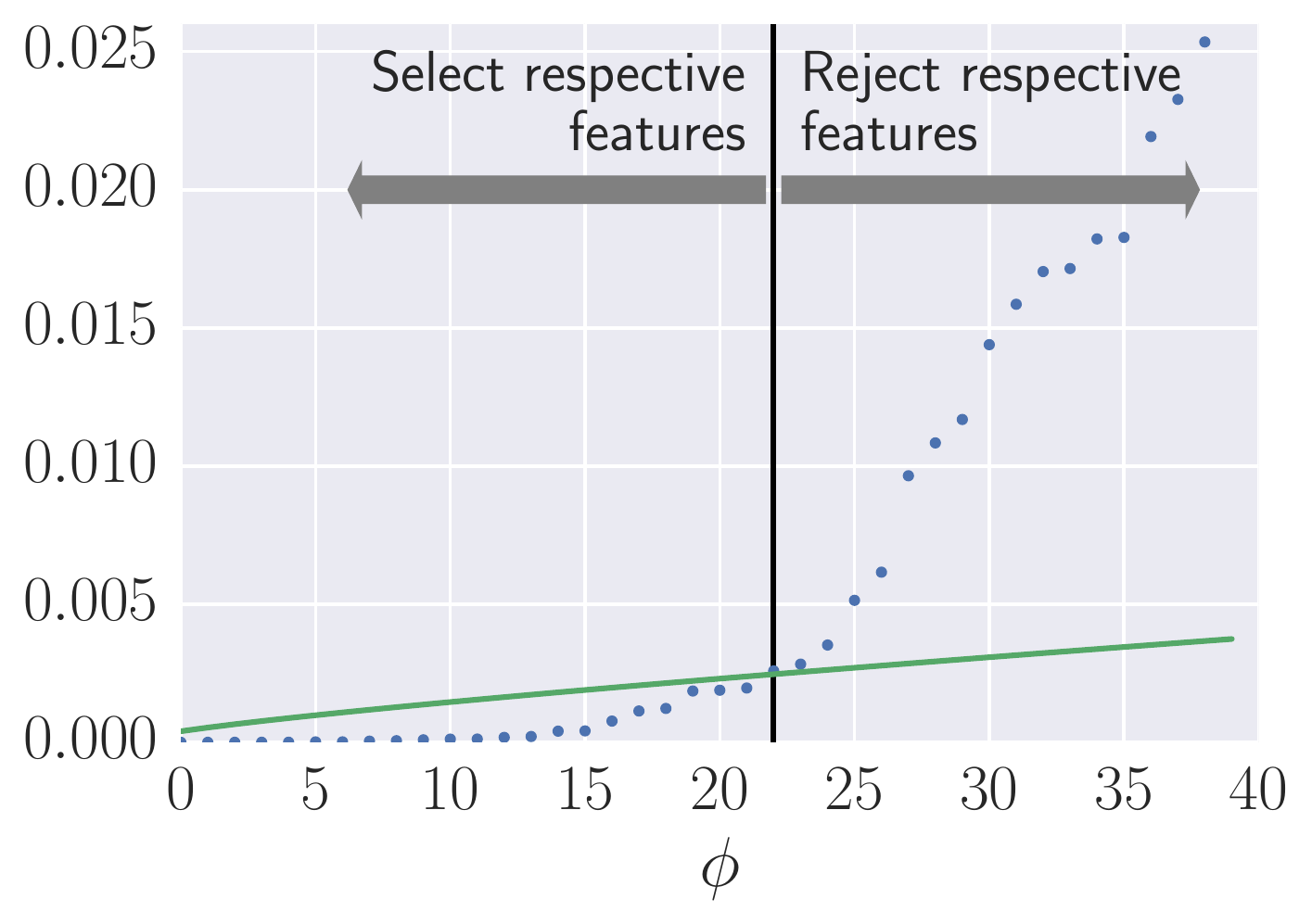}}
\caption{The Benjamini-Yekutieli procedure for a sample of simulated p-Values of 250 individual feature significance tests. 
The rejection line aims to control a FER level $q$ of 10\%.}
\label{fig:by_fdr}
\end{figure}

In the context of time series feature extraction, a wrongly added feature is a feature $X_\phi$ for which the null hypothesis $H^\phi_0$ has been rejected by the respective feature significance test, even though $H^\phi_0$ is true.
The risk of such a false positive result is by construction of the hypothesis tests only controlled for individual features.
However, when comparing multiple hypotheses and features simultaneously, errors in the inference tend to accumulate \citep{curran-everett_multiple_2000}.
In multiple testing, the expected proportion of erroneous rejections among all rejections is called false discovery rate (FDR).

The FDR as a measure of the accumulated statistical error was suggested by \citet{benjamini_controlling_1995}.
Later the non-parametric Benjamini-Yekutieli procedure was proposed.
Based on the p-values it tells which hypotheses to reject while still controlling the FDR under any dependency structure between those hypotheses \citep{benjamini_control_2001}.
It will be the last component of our filtered feature extraction algorithm.

The procedure searches for the first intersection between the ordered sequence of p-values $p_{(\phi)}$ (blue dots in Fig.~\ref{fig:by_fdr}) with a linear sequence (green lines in Fig.~\ref{fig:by_fdr})
\begin{equation}
  \label{eq:linear_sequence}
 r_{\phi}=\frac{\phi q}{n_\phi \sum_{\mu=1}^\phi \frac{1}{\mu}}.
\end{equation}
Here, $n_\phi$ is the number of all tested null hypotheses and $q$ is the FDR level that the procedure controls.
It will reject all hypotheses belonging to p-values, which have a lower value than the p-value at the intersection (Fig.~\ref{fig:by_fdr_focus}).


By deploying the Benjamini-Yekutieli procedure, a global inference error measure, the False Extraction Rate (FER) is controlled:
\begin{equation}
\label{eqn:fer}
\text{FER} = \mathbb{E} \left [ \frac{\text{number of irrelevant extracted features}}{\text{number of all extracted features}} \right ]
\end{equation}
The \alg{FRESH} algorithm controls the FER for all distributions of features and target as well as for every dependency structure asymptotically.


\subsection{The proposed feature extraction algorithm}
\label{sec:FRESH}
We combine the components from Sec.~\ref{sec:subsec_mapping},\ref{sec:subsec_featfilt} and \ref{sec:subsec_multitest} to propose \textbf{F}eatu\textbf{R}e \textbf{E}xtraction based on \textbf{S}calable \textbf{H}ypothesis tests (\alg{FRESH}) for parameter $q \in [0, 1]$, given by the following three steps:

\begin{enumerate}
\item
Perform a set of $n_\phi$ univariate feature mappings as introduced in Sec.~\ref{sec:subsec_mapping} on $m \cdot n$ different time series to create the feature vectors $X_\phi$ with $\phi=1,\ldots,n_\phi$.

\item
For each generated feature vector $X_1,\ldots,X_{n_\phi}$ perform exactly one hypothesis test for the hypothesis $H^\phi_0$ from Eq.~\eqref{eqn_glb_hypo}.
To do so, take the corresponding feature significance test from Sec.~\ref{sec:subsec_featfilt}.
Calculate the p-values $p_1,\dots, p_{n_\phi}$ of the tests. 

\item
Perform the Benjamini-Yekutieli procedure under correction for dependent hypotheses  \citep{benjamini_control_2001} for a FDR level of $q$ on the collected p-values $p_1,\dots, p_{n_\phi}$ in order to decide which null hypothesis $H_0^\phi$ to be rejected (c.f. Sec.~\ref{sec:subsec_multitest}).
Only return features vectures for which the respective hypothesis $H_0^\phi$ was rejected by the procedure.
\end{enumerate}
So, the \alg{FRESH} algorithm extracts all features in step $1.$ which are then individually investigated by the hypothesis tests in step $2.$ and finally in step $3.$ the decision about the extracted features is made. 




\subsection{Variants of \alg{FRESH}}
\label{sub:variants_FRESH}

A problem of filter methods such as the one we utilize in steps $2$ and $3$ of \alg{FRESH} is the redundancy in the feature selection.
As long as features are considered associated with the target, they will all be selected by the filter even though many of them are highly correlated to each other \citep{kira_practical_1992}.
For example the vectors for $\median$ and $\mean$ are highly correlated in the absence of outliers in the time series, and therefore both features will have similar p-values.
Hence, we expect \alg{FRESH} to either select or drop both $\median$ and $\mean$ at the same time.
To avoid generating a group of highly correlated features we propose to add another step to \alg{FRESH}:
\begin{itemize}
\item[$*.$] 
Normalize the features and perform a principal component analysis (PCA).
Keep the principal components with highest eigenvalue describing $p$ percent of the variance.
\end{itemize}
This step will reduce the number of features and 
the obtained principal components are de-correlated, orthogonal variables \citep{hotelling1933analysis}.

One could perform step $*$ between steps $1$ and $2$ of \alg{FRESH} to get rid of the correlations between the created variables early.
Then the feature significance tests in step $2$ of \alg{FRESH} will take principal components instead of the original features as input.
We will denote this variant of \alg{FRESH} as $\alg{FRESH\_PCAb}$(efore).
Also, one could perform step $*$ after the \alg{FRESH} algorithm, directly after step $3$.
This means that the PCA will only process those features, which are found relevant by the \alg{FRESH} algorithm instead of processing all features.
This variant of \alg{FRESH} is called $\alg{FRESH\_PCAa}$(fter).

Finally, due to the selection of hypothesis tests in Sec.~\ref{sec:subsec_featfilt}, \alg{FRESH} and its variants are only suitable for binary classification or regression problems.
However, by selecting suitable tests for step \emph{2.} of the algorithm, we can extend the algorithm to multi-classification problems.

\subsection{Contribution of this work}

This work makes a contributions to both the field of time series classification as well as regression of exogenous variables from time series. 
The presented feature filtering process is based on already well investigated statistical techniques such as the Kendall rank test; hence, the novelty of the \alg{FRESH} algorithm does not lay in the individual components but in the selection and combination of suitable univariate hypothesis tests (c.f. Sec.~\ref{sec:subsec_featfilt}) with a multiple testing procedure (c.f. Sec.~\ref{sec:subsec_multitest}).
This selection was made while having Big Data applications in mind, so one goal was to make \alg{FRESH} highly scalable (e.g. with respect to length, number of time series or number of extracted features); in Sec.~\ref{sec:discussion} we will discuss the distributed nature of \alg{FRESH} and further implications.

\section{Evaluation}
\label{Sec:evaluation}
In the following presented simulations, the performance of \alg{FRESH}, its two variants from Sec.~\ref{sub:variants_FRESH} and other time series feature extraction methods are compared. 
We are interested if \alg{FRESH} is able to automatically extract meaningful features and how long it takes to extract such features.
Those aspects refer to both the benefit (c.f. Sec.~\ref{sec:benefit}) and the cost dimension (c.f. Sec.~\ref{sec:costs}) of the algorithm.

During evaluation, all feature extraction methods operate on the same feature mappings (c.f. \ref{list_of_feature_mappings}), the differences lay only in the used feature selection process.
So we extract the same features but let them get selected by different approaches.
Further, the same partitions are used during folding, allowing a fair comparison between different approaches.

\subsection{Setup}

In the simulations, several feature based approaches and the shape-based classifier \alg{DTW\_NN} \citep{wang_characteristicbased_2006}, a nearest neighbor search under the Dynamic Time Warping distance, are compared.
As discussed, we extracted 111 features for every type of time series, which are then filtered by five different feature selection approaches, the first three being the introduced \alg{FRESH}, \alg{FRESH\_PCAb} and \alg{FRESH\_PCAa} (c.f. Sec~\ref{sec:FRESH} and \ref{sub:variants_FRESH}).
Further, the random tree based \alg{Boruta} feature selection algorithm \citep{kursa_all_2011} and a forward selection with a linear discriminant analysis classifier \cite{fulcher_highly_2014}, denoted by \alg{LDA}, are considered.
\alg{FRESH} is parameterized with $q=10\%$ (c.f. Eq.~\eqref{eq:linear_sequence}).
Its variants, which apply a PCA, are keeping those principal components that explain $p=95\%$ of the variance (c.f. Sec.~\ref{sub:variants_FRESH}).
Lastly, the filtering of the features by \alg{Boruta} will be based on 50 Random Forest Ensembles, each containing 10 Decision Trees.

The different extraction methods and \alg{DTW\_NN} were picked for the following reasons:
\alg{DTW\_NN} is reported to reach the highest accuracy rates among other time series classifiers \cite{geler2016comparison}, \alg{LDA} was the first proposed algorithm to automatically extract features from time series \citep{fulcher_highly_2014} and \alg{Boruta} is a promising feature selection algorithm that incorporates interactions between features, in contrast to \alg{FRESH} and its variants that evaluate features individually.

The UCR time series archive \citep{chen_ucr_2015} is a widespread benchmark environment for the time series classification community. 
We picked those 31 time series datasets from the archive that state a binary classification problem.
Also, in order to compare the runtime of the different methods, synthetic time series of flexible length and sample number belonging to two classes are generated by simulating the stochastic dynamics of a dissipative soliton \cite[p. 164]{liehr_dissipative_2013}.

The third and last data source is from the production of \iPRODICTproduct{}, extracted during the German research project \iPRODICT{}.
This project demonstrates a typical application of industrial time series analysis, aiming to predict the passing or failing of product specification testings based on temporally annotated data.
The steel producer can adapt his business processes during or directly after the production with such forecasts.
Without them, he has to wait for the results of the specifications tests for which the steel billets need to cool down, a process that can take several hours or days \cite{Christ2016_CEP}.
Therefore, the predictions will enable faster business process adaptions and save costs by increasing the agility and efficiency of the production.
This dataset has 26 univariate meta-variables forming the baseline feature set extended by 20 different sensor time series having up to 44 data points for each sample.
The dataset contains 5000 samples of two classes \emph{"broken"} and \emph{"not broken"}. 

\subsection{The benefit - achieved Accuracy}
\label{sec:benefit}

For datasets from both the UCR time series repository as well as the \iPRODICT{} project, the underlying structure and therefore the relevant features are unknown.
We cannot compare the different methods on their ability to extract meaningful features because we do not know which features are meaningful and which not.
Also, we cannot compare the extracted features to shape-based classifiers such as \alg{DTW\_NN}, who do not extract features.
Therefore, we evaluate the performance of the feature extraction algorithms by comparing the performance of a classification algorithm on the extracted features.
Hereby, we assume that more meaningful features will result in a better classification result.


So, we need to select classification algorithms that can be trained on the (filtered) feature matrices.
In \cite{fernandez-delgado_we_2014}, \citeauthor{fernandez-delgado_we_2014} evaluate a range of 179 classifiers from 17 families on 121 different datasets representing a wide range of different applications.
In their evaluation, the Random Forest Classifiers formed the best performing group.
Hence, in our evaluation, we will also evaluate our algorithms on a Random Forest Classifier, denoted by \alg{rfc}. 
Further, we consider an AdaBoost Classifier \cite{freund1995desicion}, denoted by \alg{ada}, due to boosting algorithms performing generally well on a wide range of problems \cite{fernandez-delgado_we_2014,alfaro2008bankruptcy,jones2015empirical} and AdaBoost being considered on of the best \emph{``out-of-the-box classifiers''} \cite{kegl2013return}. 
The hyperparameters for those methods are not optimized to get an unbiased view on the meaningfulness of the extracted features, instead the default values from the python package scikit-learn version $0.18.1$ were used \citep{pedregosa_scikitlearn_2011}.

Combining the five feature selection algorithms with the two classifiers will results in 10 different feature based approaches. 
We append the name of the classifier as postfix to the approach name (e.g. \alg{LDA\_ada} denotes a filtering of the features by the \alg{LDA} algorithm with subsequent classification by \alg{ada}).
This gives us the pipelines
\alg{FRESH\_rfc}, \alg{FRESH\_ada},
\alg{Boruta\_rfc}, \alg{Boruta\_ada}, 
\alg{LDA\_rfc}, \alg{LDA\_ada}, 
\alg{FRESH\_PCAa\_rfc}, \alg{FRESH\_PCAa\_ada},
\alg{FRESH\_PCAb\_rfc} and \alg{FRESH\_PCAb\_ada}.
Further, we apply the two considered classifiers on the unfiltered feature matrix, then denoted by \alg{rfc} and \alg{ada}.
Finally, we included \alg{trivial}; a benchmark algorithm that disregards any relationship between time series and class label by always predicting the majority class on the training dataset.
Algorithms that learn an informative mapping between time series and exogenous target variables will outperform this simple benchmark.

\begin{figure}[tbp]
 \includegraphics[width=\linewidth]{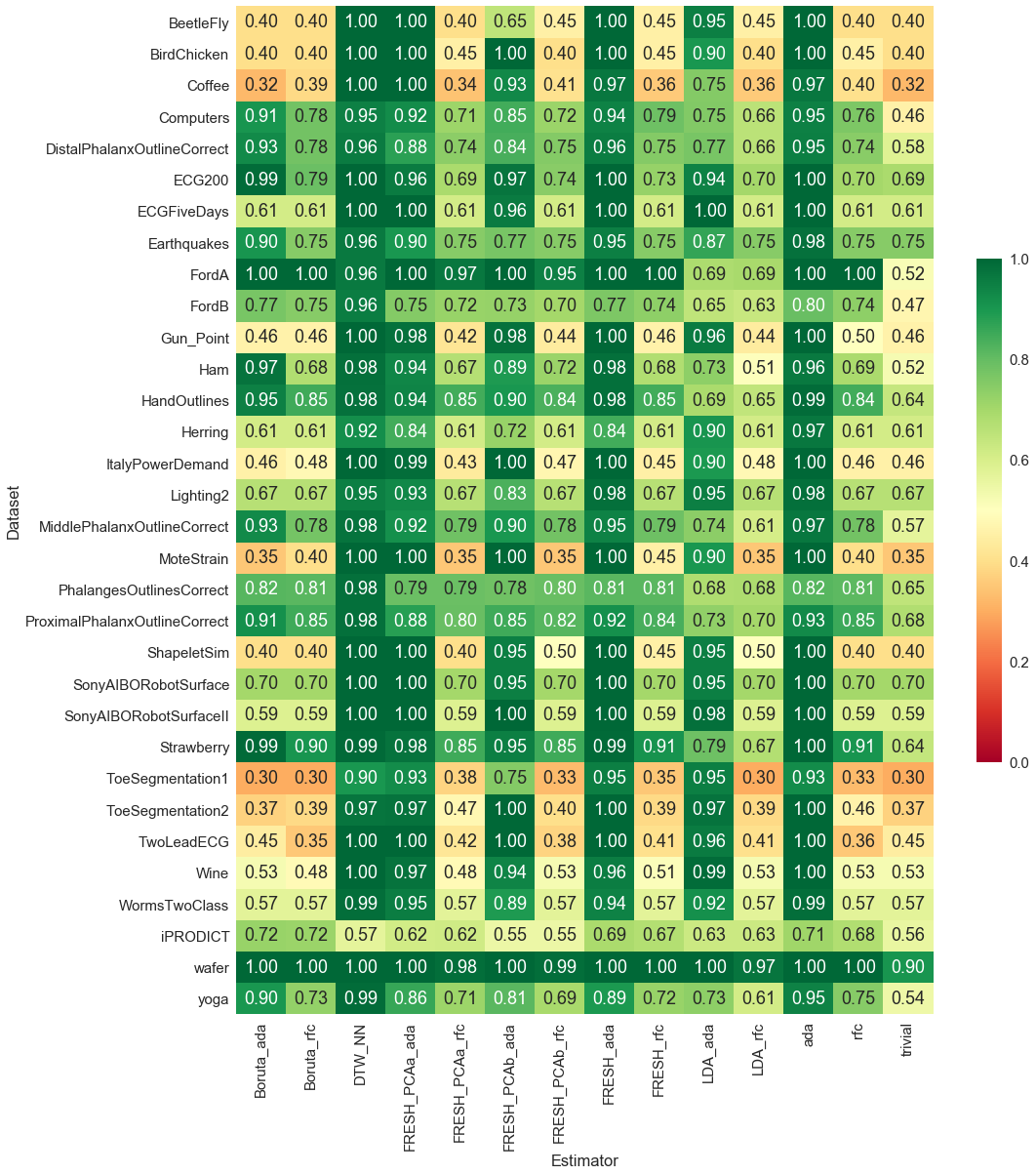}
 \caption{\textbf{Average value of the accuracy:}
Accuracy of the different feature extraction methods and \alg{DTW\_NN} on the 31 two-class datasets from the UCR time series archive as well as the data from the \iPRODICT{} research project.
The reported accuracy was averaged over a 10 fold cross validation, where every approach had access to the same folds to ensure a fair comparison.}
\label{fig:heatmap_accuracy}
\end{figure}

\begin{figure}[tbp]
 \includegraphics[width=\linewidth]{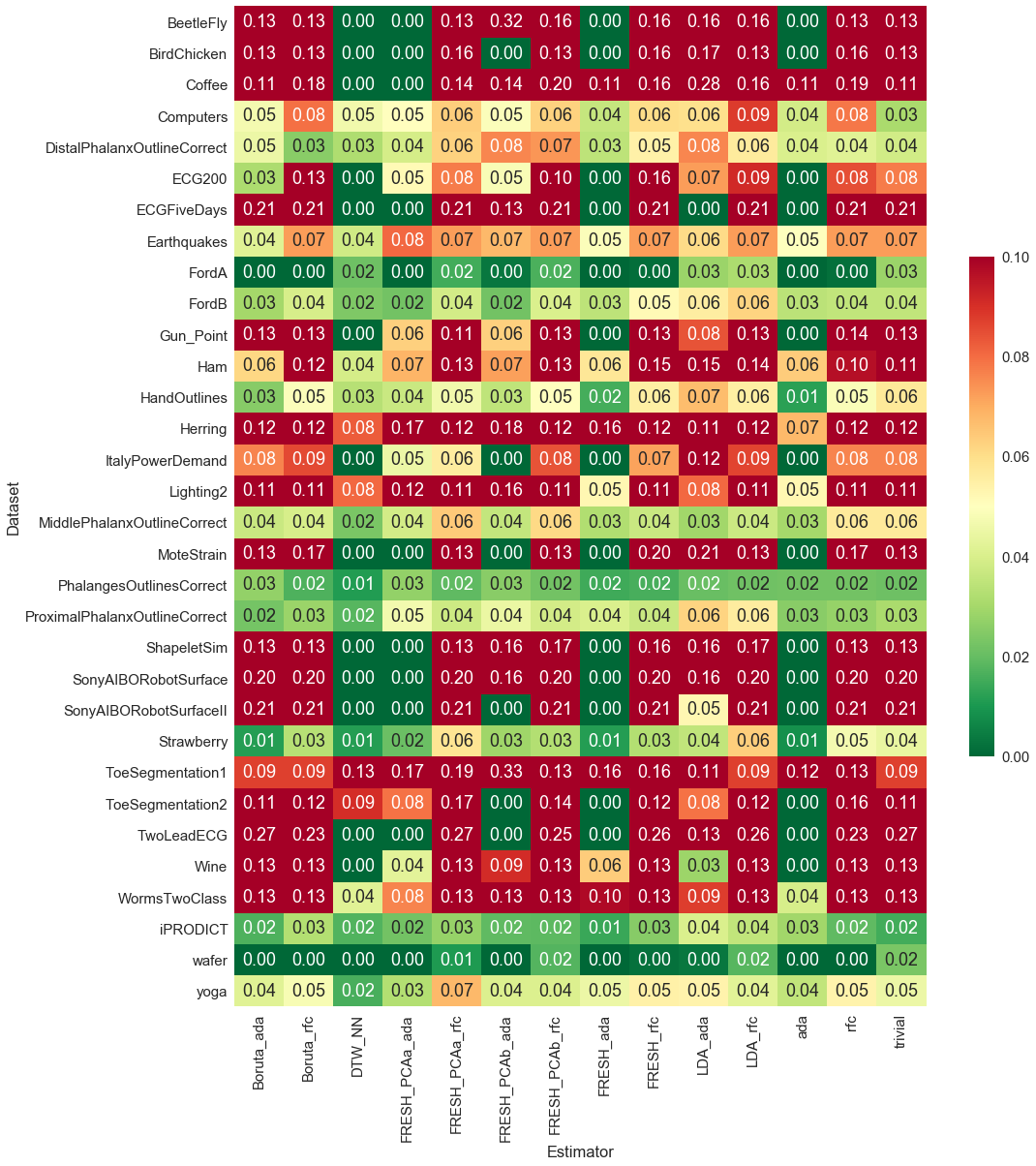}
 \caption{\textbf{Standard deviation of the accuracy:}
This heatmap shows the standard deviation of the accuracy metric from Fig.~\ref{fig:heatmap_accuracy} for different combinations of approaches and datasets during a 10 fold cross validation.
}
\label{fig:heatmap_accuracy_std}
\end{figure}

For every dataset, all available samples will be used to perform the feature extraction itself.
Then, the feature selection steps and classifiers are trained using a 10 fold cross validation scheme.
The average accuracies of this evaluation are contained as a heatmap in Fig.~\ref{fig:heatmap_accuracy}.
The value in a singular cell is the calculated accuracy for a singular approach-dataset combination, averaged over the 10 folds.
Furthermore, Fig.~\ref{fig:heatmap_accuracy_std} reports the standard deviation of the accuracy metric over the 10 folds. 

The best performing algorithms were \alg{ada} and \alg{DTW\_NN}, achieving a  highest mean accuracy on 21 out of the 32 datasets, with all 21 originating from the UCR time series archive.
In addition, both approaches showed similar variance in the results; when comparing the standard deviation that is contained in Fig.~\ref{fig:heatmap_accuracy_std}, the \alg{DTW\_NN} has a higher standard deviation than \alg{ada} for the achieved accuracy on 9 datasets while vice-versa the standard deviation of the accuracy by \alg{ada} is higher on 10 datasets.
This shows that \alg{ada} and \alg{DTW\_NN} were both comparable with respect to achieved accuracy.

On the dataset from the \iPRODICT{} research project, the Boruta based approaches \alg{Boruta\_ada} and \alg{Boruta\_rfc} showed the highest mean accuracies of 72\%, with plain AdaBoost \alg{ada} coming close behind with a mean accuracy of 71\%.
The \alg{DTW\_NN} algorithm was only trained on 1 out of 20 sensor time series for the \iPRODICT{} data and was not using any univariate variables (we trained \alg{DTW\_NN} for all 20 time series and then picked the one time series with the best results) which explains its low mean accuracy of 57\%.

Regarding the trivial benchmark, only \alg{DTW\_NN}, \alg{LDA\_ada}, \alg{ada}, \alg{FRESH\_ada}, \alg{FRESH\_PCAa\_ada} and \alg{FRESH\_PCAb\_ada} were able to beat it on every dataset.
Every Random Forest based approach reported a mean accuracy that was lower than the trivial one on at least one dataset.
For example, on the Wine dataset \alg{Boruta\_rfc}, \alg{FRESH\_rfc} and \alg{FRESH\_PCAb\_rfc} were performing worse than the trivial benchmark.
It seems that the Random Forest approach is not as robust as AdaBoost.

But, overall, the reported accuracy and the good performance of \alg{ada} show that feature based approaches are competitive to shape-based approaches such as \alg{DTW\_NN} with respect to prediction accuracy.
Our simulations show that an out-of-box classifier without any hyperparameter optimization and a generic set of time series features (c.f. \ref{list_of_feature_mappings}) is able to be on par with a state-of-the-art shape-based approach.

Further, we were interested in the change of accuracy when deploying different feature selection techniques.
For \alg{ada}, the filtering by \alg{FRESH}, \alg{FRESH\_PCAa},  \alg{FRESH\_PCAb}, \alg{Boruta} and \alg{LDA}, improved or not changed the mean accuracy in 19, 12, 7, 2 and 5 out of 32 datasets, respectively.
Regarding the second classifier \alg{rfc}, this happened on 22, 15, 17, 13 and 20 out of the 32 datasets.
The \alg{FRESH} filter even increased the mean accuracy of \alg{rfc} on 14 out of 32, so on nearly half of the datasets.  
This shows that the selection of features by \alg{FRESH} had the best chance among the considered feature selection techniques of not worsening the classifiers accuracy.
However, this also shows that for some data sets, the feature filtering removed important information by dropping relevant features.
(e.g. for \alg{rfc}, the mean accuracy decreased in 13 out of 32 datasets when filtering the features by \alg{FRESH}).

As always, there is no best algorithm for all kind of applications.
As expected, the \alg{DTW\_NN} approach had one of the highest accuracies in our evaluation.
However, we were surprised to see that the \alg{ada} approach without any filtering of the features reached such a competitive accuracy.
Regarding the feature filtering, we showed that \alg{FRESH} for a Random Forest Classifier on two thirds of datasets did not worsen the accuracy of the final classification algorithms while reducing the number of considered features. 
In comparison to \alg{Boruta} and \alg{LDA}, it was most often able to improve the performance of the Random Forest Classifiers.
For AdaBoost it seems, that for a low number of features (111 on the UCR datasets and 2220 on the \iPRODICT{} data) and estimators (the scikit-learn default are 50 decision trees in the ensemble), the feature filtering is not necessary as the algorithm not yet tends to overfit.

\subsection{The costs - needed Runtime}
\label{sec:costs}

The runtime of algorithms highly depend on the efficiency of the implementation and the characteristics of the used hardware.
By using a slight more inefficient implementation to compare your own algorithm against, you can skew the picture in your favor.
Therefore we decided to only inspect asymptotic runtimes and not discuss absolute runtimes in the evaluation.

\begin{figure}[tbp]
   \subfigure[Number of samples fixed to 1000]{
   \includegraphics[width=0.48\linewidth]{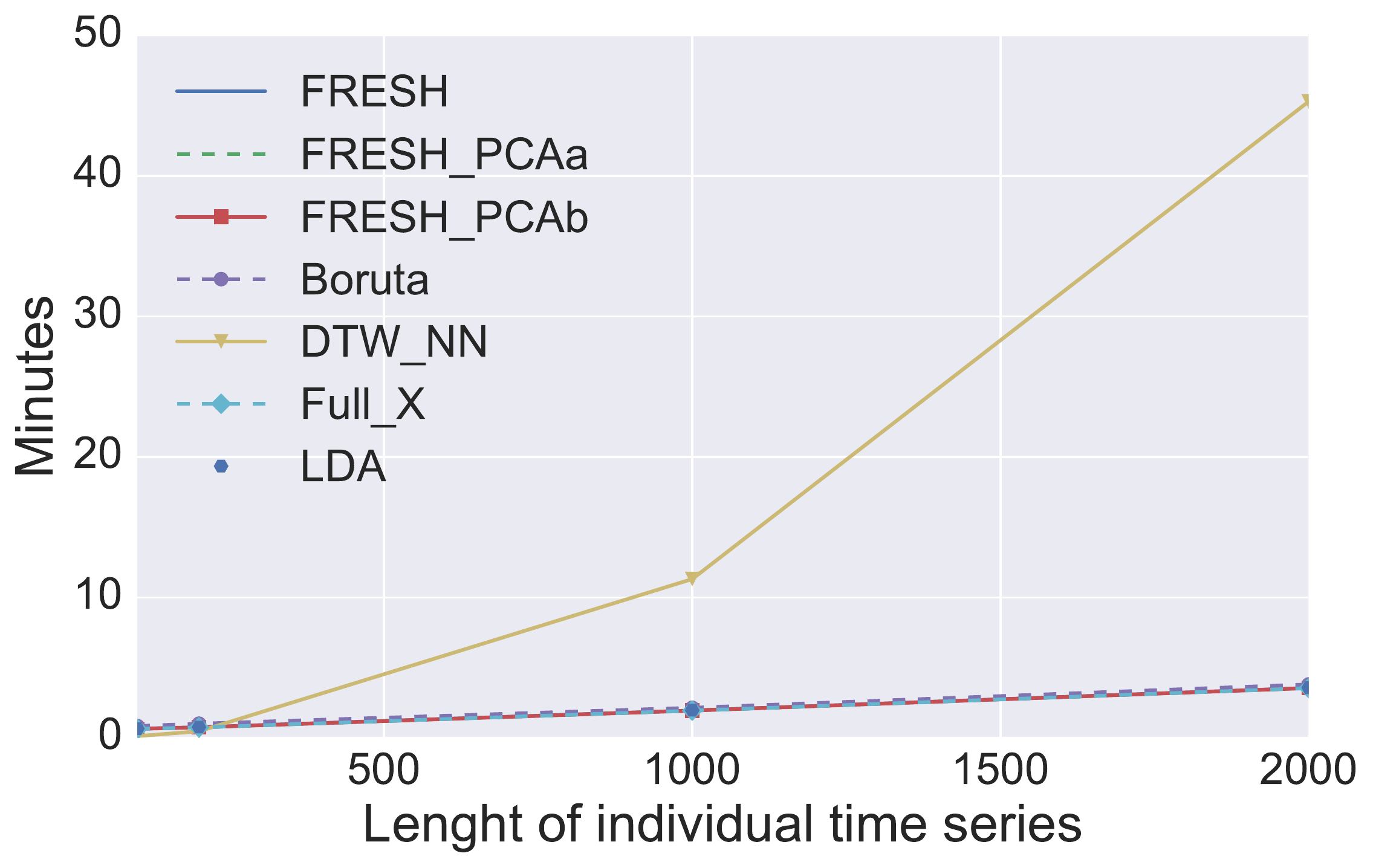}
   \label{fig:runtime_length_ts_n_samples_1}}
   \subfigure[Length of time series fixed to 1000]{\includegraphics[width=0.48\linewidth]{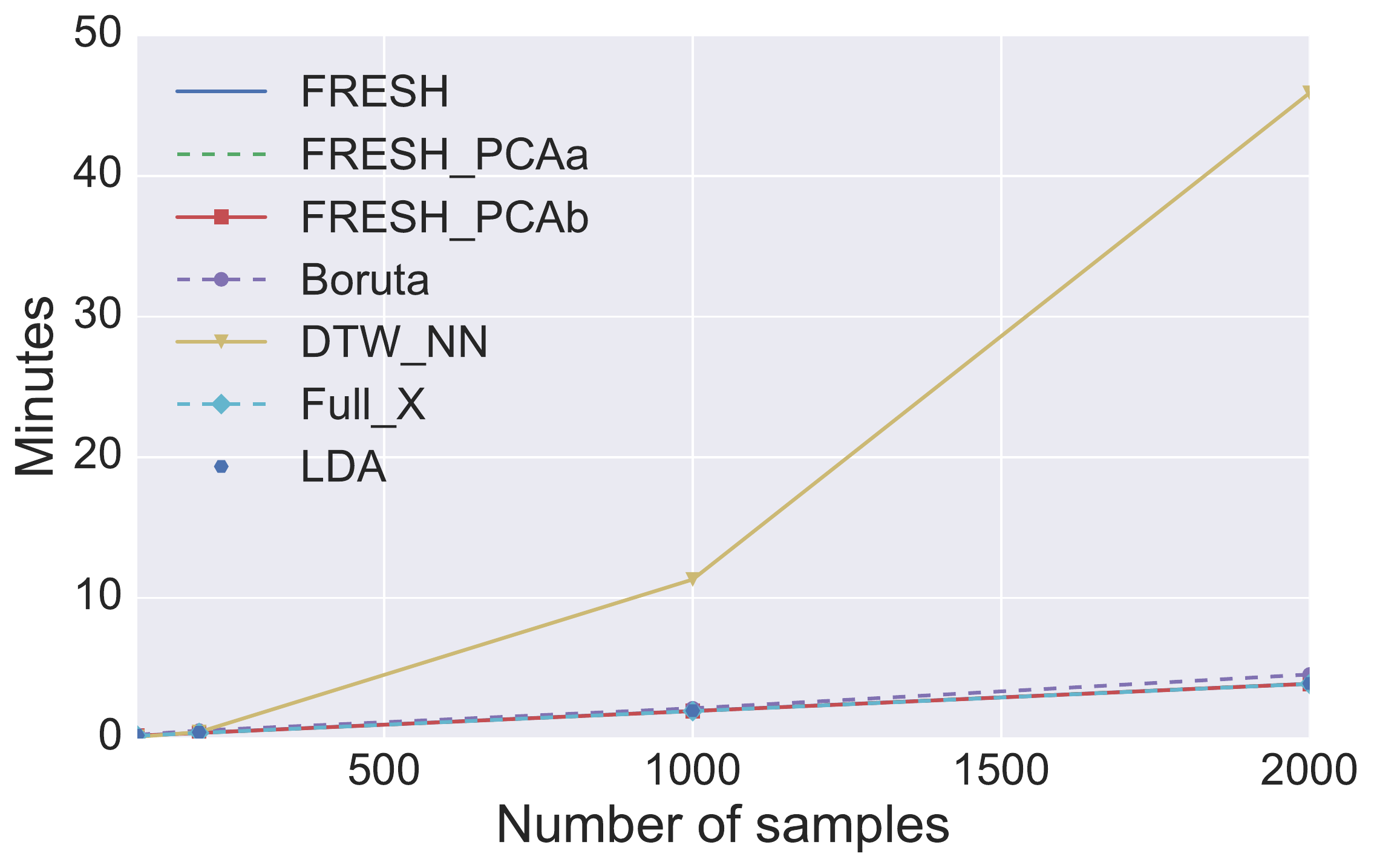}
   \label{fig:runtime_length_ts_n_samples_2}}
\caption{\textbf{Average pipeline runtime} of time series classification concerning the nonlinear dynamics of a dissipative soliton \cite[p. 164]{liehr_dissipative_2013}. 
The reported durations are the summed runtimes of feature extraction, feature filtering and predicting in the case of a feature based approaches, for \alg{DTW\_NN} it just denotes the fitting and predicting.
For the feature based approaches, the fitting and predicting runtimes of a group of five classificators have been averaged: a layer neural network/perceptron, a logistic regression model, a Support Vector Machine, a Random Forest Classifier and an AdaBoost Classier.
\alg{Full\_X} denotes the pipeline without any feature filtering.
The curves of all methods except \alg{DTW\_NN} lay on top of each other.
We can observe that all feature based approaches scale linearly with the number of samples and length of time series, in contrast to \alg{DTW\_NN}.}
\label{fig:runtime_length_ts_n_samples}
\end{figure}

We are interested in the feature extraction method's ability to scale with an increasing number of feature mappings, time series length and device numbers.
As expected, Fig.~\ref{fig:runtime_length_ts_n_samples} shows that all considered feature extraction methods -- in contrast to \alg{DTW\_NN} -- scale linearly with an increasing length of the time series or increasing number of samples.
This is due to the considered feature mapping having a linear runtime with respect to the length of the time series.

\begin{figure}[tbp]
   \subfigure[All methods]{
   \includegraphics[width=0.48\linewidth]{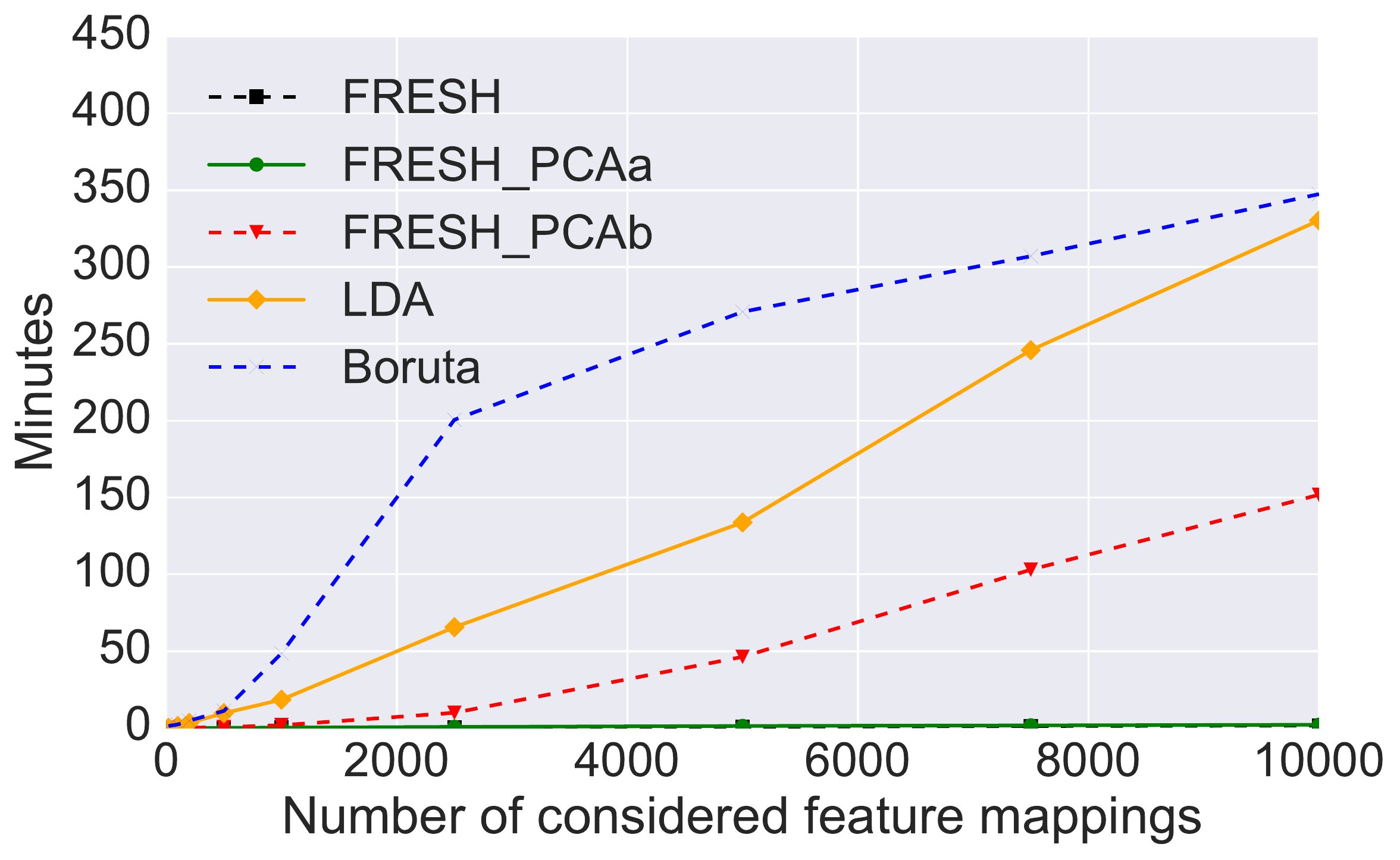}
   \label{fig:runtime_fsa_nfeatures_1}
   }
   \subfigure[Only $FRESH$ and variants]{
     \includegraphics[width=0.48\linewidth]{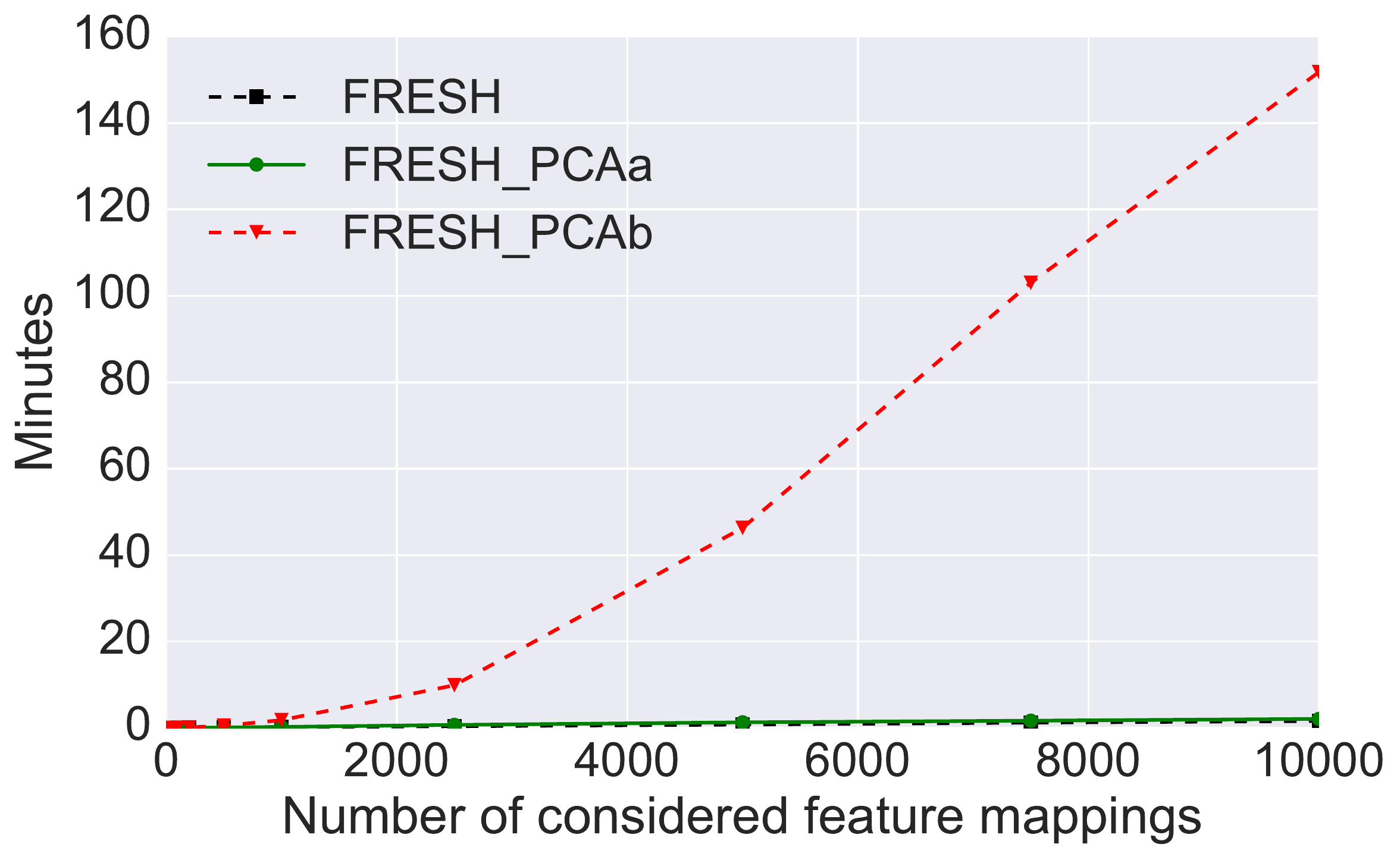}
     \label{fig:runtime_fsa_nfeatures_2}
     }
\caption{\textbf{Average feature extraction runtime} during ten feature selection runs for 10,000 samples of a variable number of feature mappings (the curves of \alg{FRESH} and \alg{FRESH\_PCAa} are overlapping).
We can observe that only \alg{FRESH} and its variants scale linearly with the number of considered features.
The red line of \alg{FRESH\_PCAb} is due to the PCA being calculated on all features while for \alg{FRESH\_PCAa} the filtered feature matrix is used}
\label{fig:runtime_fsa_nfeatures}
\end{figure}

However, Fig.~\ref{fig:runtime_fsa_nfeatures} shows that, among the feature based approaches, only \alg{FRESH} and \alg{FRESH\_PCAa} scale linear with an increasing number of features (e.g. due to more devices, feature mappings or types of time series).
This makes \alg{FRESH} and \alg{FRESH\_PCAa} suitable to filter huge amounts of time series features in Big Data applications.

%
%
%
%
%

\subsection{Selected features}



\begin{figure}[t]
\center
 \includegraphics[width=0.8\linewidth]{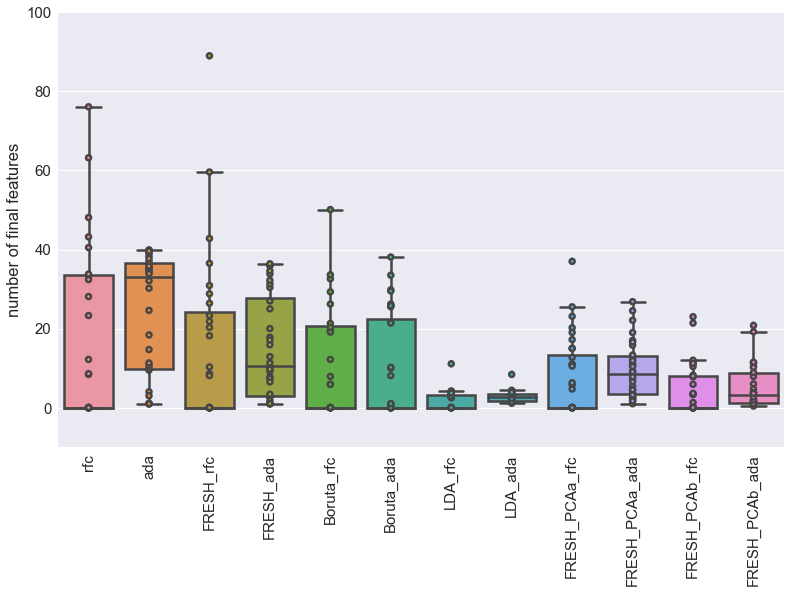}
 \caption{\textbf{Relevant features in the final classificator:}
 This boxplot shows the number of features that were used by the final Adaboost \alg{ada} and Random Forest Classificators \alg{rfc}. 
Every data point is corresponds to the result of one of the 10 folds for each of the 32 datasets for one of the two classifier. 
}
\label{fig:boxplot_n_features}
\end{figure}

We were interested in how many features the final classification algorithm was based on.
Both \alg{ada} and \alg{rfc} are able to select features for the splits inside their decision trees.
This means that they are able to internally select features.
Fig.~\ref{fig:boxplot_n_features} contains boxplots of the number of final features that were picked by the two classification algorithms.

We can observe that all feature selection methods, so \alg{Boruta}, \alg{LDA} and the \alg{FRESH} variants, were able to reduce the number of selected features for both final classifiers \alg{rfc} and \alg{ada}.
The filtering by \alg{LDA} resulted in the lowest number of relevant features.
This is due to \alg{LDA} selecting features in a stepwise nature. 
The algorithm will start with no features and only add features if they improve the linear classification rate; this greedy, forward selection process can run into local minima and will often result in a small sets of considered features.

As expected, the filtering by both \alg{FRESH\_PCAa} and \alg{FRESH\_PCAb} will result in a lower number of relevant features in the final classificators than with \alg{FRESH}.
This due to the filtering by the PCA step that works as a second filter step.
Out of the three variants, \alg{FRESH\_PCAb} was able to reduce the number of features in the final classification algorithm the most.

\subsection{Resume}
We proposed \alg{FRESH} as a highly scalable feature extraction algorithm.
Our simulations showed that, in contrast to other considered methods, \alg{FRESH} is able to scale with the number of feature mappings and samples as well as with the amount of different types and length of the time series.
While doing so, it is extracting meaningful features as demonstrated by competitive accuracies.

The most compelling feature selection method was \alg{FRESH}, without any PCA step.
When comparing the change in accuracy to the base classificator, in 14 out of 32 data sets the accuracy of a Random Forest Classifier increased while in 8 it stayed the same.
Also, for the AdaBoost Classifier, the \alg{FRESH} filtering was most often not worsening the performance with respect to accuracy.

The relative bad performance of $\alg{FRESH\_PCAb}$ seems to originate in the PCA step selecting features only based on their ability to explain the variance in the input variables and not in their significance to predict the target variable.
By this, relevant information for the classification or regression task can get lost.
Also, it was more effective for the Random Forest Classifier than for the AdaBoost approach.

On the other hand, the combination of \alg{FRESH} with a subsequent PCA filtering to reduce the number of redundant and highly correlated features, denoted as \alg{FRESH\_PCAa} showed similar performance for both classifiers, its reduced number of features resulted in a lower accuracy on 20 respectively 17 datasets.

\section{Discussion}
\label{sec:discussion}
\subsection{FRESH assists the acquisition of domain knowledge}


It is common knowledge that the quality of feature engineering is a crucial success factor for supervised machine learning in general \cite[p. 82]{domingos_few_2012} and for time series analysis in particular \citep{timmer_characteristics_1993}.
But comprehensive domain knowledge is needed in order to perform high quality feature engineering.
Contrarily, it is quite common for machine learning projects that data scientists start with limited domain knowledge and improve their process understanding while continuously discussing their models with domain experts.
This is basically the reason, why dedicated time series models are very hard to build from scratch. 

Our experience with data science projects in the context of IoT and Industry 4.0 applications \citePoster{} showed that it is very important to identify relevant time series features in an early stage of the project in order to engineer more specialized features in discussions with domain experts.
The \alg{FRESH} algorithm supports this approach by applying a huge variety of established time series feature mappings to different types of time series and meta-information simultaneously and identifies relevant features in a robust manner.

We observe that features extracted by \alg{FRESH} contribute to a deeper understanding of the investigated problem, because each feature is intrinsically related to a distinct property of the investigated system and its dynamics.
This fosters the interpretation of the extracted features by domain experts and allows for the engineering of more complex, domain specific features \citePoster{} including dedicated time series models, such that their predictions in return might become a future feature mapping for \alg{FRESH}.




\subsection{FRESH is operational} 

We have already mentioned that \alg{FRESH} has been developed in the course of IoT and Industry 4.0 projects \citePoster{}. 
Especially for predictive maintenance applications with limited numbers of samples and high level of noise in e.g. sensor readings, it has been proven as crucial to filter irrelevant features in order to prevent overfitting. 
To ensure a robust and scalable filtering, we consider each feature importance individually.
This causes several implications:
\begin{itemize}
\item 
\alg{FRESH} is robust in the sense of classical statistics,  because the hypothesis tests and the Benjamini-Yekutieli procedure do not make any assumptions about the probability distribution or dependence structure between the features.
Here, robustness refers to the insensitivity of the estimator to outliers or violations in underlying assumptions \citep{john_robustness_2013}.
\item 
\alg{FRESH} is not considering the meaningfulness of interactions between features by design.
Hence, in its discussed form it will not find meaningful feature combinations such as chessboard variables \cite[Fig.~3a]{guyon_introduction_2003}.
However, in our evaluation process the feature selection algorithm Boruta, which considers feature interactions, was not able to beat the performance of \alg{FRESH}.
Further, it is possible for \alg{FRESH} to incorporate combinations of features and pre-defined interactions as new features themselves. 
\item
\alg{FRESH} is scalable due to the parallelity of the feature calculation and hypothesis tests (see the two topmost tiers in Fig.~\ref{fig:ffe_in_detail}) and can be trivially parallelized and distributed over several computational units. 
Because we only deploy stateless features, so the calculation of each feature for each device does not depend on the other features, it is trivial to parallelize the task of feature calculation both horizontally (over different features) and vertically (over different entities).
In addition, the feature filter process has low computational costs compared to feature calculation and significance testing.
Therefore, \alg{FRESH} scales linearly with the number of extracted features, length of the time series, and number of considered time series making it a perfect fit for (industrial) Big Data applications.


\item 
A side effect of ensuring robustness and testing features individually is that \alg{FRESH} tends to extract highly correlated features, which could result in poor classification performance.  
We propose to combine \alg{FRESH} with a subsequent PCA, which has been discussed as \alg{FRESH\_PCAa} in Sec.~\ref{sub:variants_FRESH} and indeed improved the performance significantly. 

\end{itemize}

\subsection{Feature selection of FRESH}


\citet{nilsson_consistent_2007} proposed to divide feature selection into two flavors: The \emph{minimal optimal problem} is finding a set consisting of all strongly relevant attributes and a subset of weakly relevant attributes such that all remaining weakly relevant attributes contain only redundant information.
The \emph{all-relevant problem} is finding all strongly and weakly relevant attributes.
The first problem is way harder than the second, even asymptotically intractable for strictly positive distributions \citep{nilsson_consistent_2007}.
Accordingly, \alg{FRESH} solves the second, easier problem as we extract every relevant feature, even though it might be a duplicate or highly correlated to another relevant feature \cite{kursa_all_2011}.

\citet{yu_feature_2003} separated feature selection algorithms into two categories, the \emph{wrapper model} and the \emph{filter model}.
While the selection of wrapper models is based on the performance of a learning algorithm on the selected set of features, filter models use general characteristics to derive a decision about which features to keep.
Filter models are further divided into feature weighting algorithms, which evaluate the goodness of features individually, and subset search algorithms, which inspect  subsets.
According to this definition, the feature selection part of \alg{FRESH} is a filter model, more precisely, a feature weighting algorithm with the weights being the p-values assigned to the corresponding features.




\alg{FRESH} contains a feature selection part on basis of hypothesis tests and the Benjamini-Yekutieli procedure, which of course can be used as a feature selection algorithm on features derived from manifold structured data such as spectra, images, videos and so on.
But, due to its systematic incorporation of scalable time series feature mappings and the proposed decomposition in computing tiers (Fig.~\ref{fig:ffe_in_detail}) it is especially applicable to the needs of mass time series feature extraction and is considered as a time series feature extraction algorithm.

For a given set of attributes, feature selection techniques decide which attributes to delete and which to keep. 
While doing so, a possible evaluation metric is the expected ratio of deleted relevant features to all deleted features.
This rate is the False Deletion Rate ($FDR$) and is formally defined by  
\begin{equation*}
\text{FDR} = \mathbb{E} \left [ \frac{\text{number of relevant but deleted features}}{\text{number of all deleted features}} \right ]
\end{equation*}
In contrast, \alg{FRESH} controls the $FER$ as defined in Eq.~\eqref{eqn:fer}. 
A control of the $FDR$ by \alg{FRESH} would need a testing of the following hypotheses
\begin{equation}
\begin{split}
\label{eq:fdr_hypo}
H_0^\phi &= \{ X_\phi \text{ and } Y \text{ are dependent} \}, \\
H_1^\phi &= \{ X_\phi \text{ and } Y \text{ are independent} \}.
\end{split}
\end{equation}
Due to the topology of the hypotheses in Eq.~\ref{eq:fdr_hypo} such a hypothesis testing is statistically not feasible.
This means that \alg{FRESH} can not be adapted to control the $FDR$.
Hence, the feature filtering of \alg{FRESH} is not suitable for feature selection jobs where the $FDR$ has to be controlled.

Finally, by applying a multiple testing algorithm, \alg{FRESH} avoids the \emph{``look-elsewhere effect''} \citep{gross_trial_2010} which is a statistically significant observation arising by chance due to the high number of tested hypotheses. 
This effect triggered a recent discussions about the use of p-values in scientific  publications \citep{wasserstein_asas_2016}.

\subsection{Related work}



There are both structural and statistical approaches to extract patterns from time series.
Many statistical approaches rely on structures that allow the usage of genetic algorithms. 
They express the feature pattern for example as a tree \citep{mierswa_automatic_2005, geurts_pattern_2001, eads_genetic_2002}.
While doing so, they aim for the best pattern and the most explaining features by alternating and optimizing the used feature mappings.
In contrast, \alg{FRESH} extracts the best fitting of a fixed set of patterns.

As an example for a structured pattern extraction, Olszewski detects six morphology types \citep{olszewski_generalized_2001}:  constant, straight, exponential, sinusoidal, triangular, and rectangular phases.
Those phases are detected by structure detectors which then output a new time series whose values stand for the identified structure.
Based on this structure a domain-independent structural pattern recognition system is utilized to substitute the original time series signal by a known pattern.
Due to its fixed patterns, \alg{FRESH} can be considered to be a structured pattern extractor.

Of course, there are other promising approaches like the combination of nearest neighbor search with Dynamic Time Warping \citep{wang_characteristicbased_2006}, which is specialized on considering an ensemble of exactly one dedicated time series type and cannot take meta-information into account.
For binary classifications it scales with $\mathcal{O}(n_t^2 \cdot m_\text{train} \cdot m_\text{test})$ \citep{penserini_comparison_2006} with $m_\text{train}$ and $m_\text{test}$ being the number of devices in the train and test set, respectively.
This approach also has the disadvantage that all data have to be transmitted to a central computing instance.

The extraction algorithm most similar to ours is presented by \citet{fulcher_highly_2014}. 
It applies a linear estimator with greedy search and a constant initial model to identify the most important features, which has been considered in this paper as LDA. 
The evaluation has shown, that \alg{FRESH} outperforms the approach of \citet{fulcher_highly_2014}.
Also, \alg{FRESH} provides a more general approach to time series feature extraction, because it is able to extract features for regression tasks and not only for classification.


\section{Summary and future work}
\label{sec:summary}
In this work, \FRESH{} (\alg{FRESH}) for time series classification and regression is introduced.
It combines well established feature extraction methods with a scalable feature selection based on non-parametric hypothesis tests and the Benjamini-Yekutieli procedure. 

\alg{FRESH} is highly parallel and suitable for distributed IoT and Industry 4.0 applications like predictive maintenance or process line optimization, because it allows to consider several different time series types per label and additionally takes meta-information into account.  
The latter has been demonstrated on basis of a \iPRODICTproduct{} process line optimization of project \iPRODICT{} \cite{Christ2016_CEP}.

Our evaluation for UCR time series classification tasks has shown that \alg{FRESH} is able to filter feature such that the performance of an AdaBoost or a Random Forest classificators are not worsened on the majority of datasets.
Interestingly, an Adaboost Classifier without any filtering of features proved to reach the highest accuracies among all feature based approaches, it was even able to beat a shape-based nearest neighbor search under a Dynamic Time Warping distance metric.

The parallel nature of \alg{FRESH} with respect to both feature extraction and filtering makes it highly applicable in situations where data is fragmented over a widespread infrastructure and computations cannot be performed on centralized infrastructure.
Due to its robustness and applicability to machine learning problems in the context of IoT and Industry 4.0, we are expecting that \alg{FRESH} will find widespread application. 

\alg{FRESH} has been developed to extract meaningful features for classification and regression tasks.
Therefore, it can be easily combined with domain specific and possibly stateful feature mappings from more specialized machine learning algorithms like e.g. Hubness-aware classifiers \cite{Buza2015_time_series}. 
The considered datasets in this work only contained binary classification variables.
We plan to investigate the performance of \alg{FRESH} for multi-class or regression targets in the future.

Also, we are planning to investigate other hypothesis tests to address feature significances.
For example, a substitute for the Kolmogorov-Smirnov is the Mann-Whitney U test which checks if the median of two variables $A$ and $B$ differs \citep{MannWhitney1947_DA}.
While the Kolmogorov-Smirnov test is sensitive to any differences in the distributions with regard to shape, spread or median, the Mann-Whitney U test is mostly sensitive to changes in the median.
We will investigate how such different hypothesis tests affect the performance of \alg{FRESH}.

As the only part of \alg{FRESH} that depends on the structure of the data are the used feature mappings, one could easily inspect the performance of the algorithm on other domains such as image or video processing.
We are curious to see further research in this area.

\section*{Acknowlegement}
The authors would like to thank Frank Kienle and Ulrich Kerzel for fruitful discussions as well as Nils Braun and Julius Neuffer for their most valuable contributions during the implementation of the \alg{tsfresh} package. 
This research was funded in part by the German Federal Ministry of Education and Research under grant number \iPRODICTgrant{} (project \iPRODICT{}).

\section*{References}

\newpage 

\section*{Vitae}

\parpic{\includegraphics[width=1in,clip,keepaspectratio]{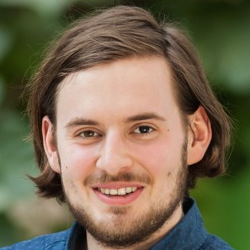}}
\noindent {\bf Maximilian Christ} received his M.S. degree in Mathematics and Statistics from Heinrich-Heine-Universit\"{a}t D\"{u}sseldorf, Germany, in 2014.
In his daily work as a Data Science Consultant at Blue Yonder GmbH he optimizes business processes through data driven decisions.
Beside his business work he is persuing a Ph.D in collaboration with University of Kaiserslautern, Germany. 
His research interest relate on how to deliver business value through Machine Learning based optimizations.
\vspace{1cm}

\parpic{\includegraphics[width=1in,clip,keepaspectratio]{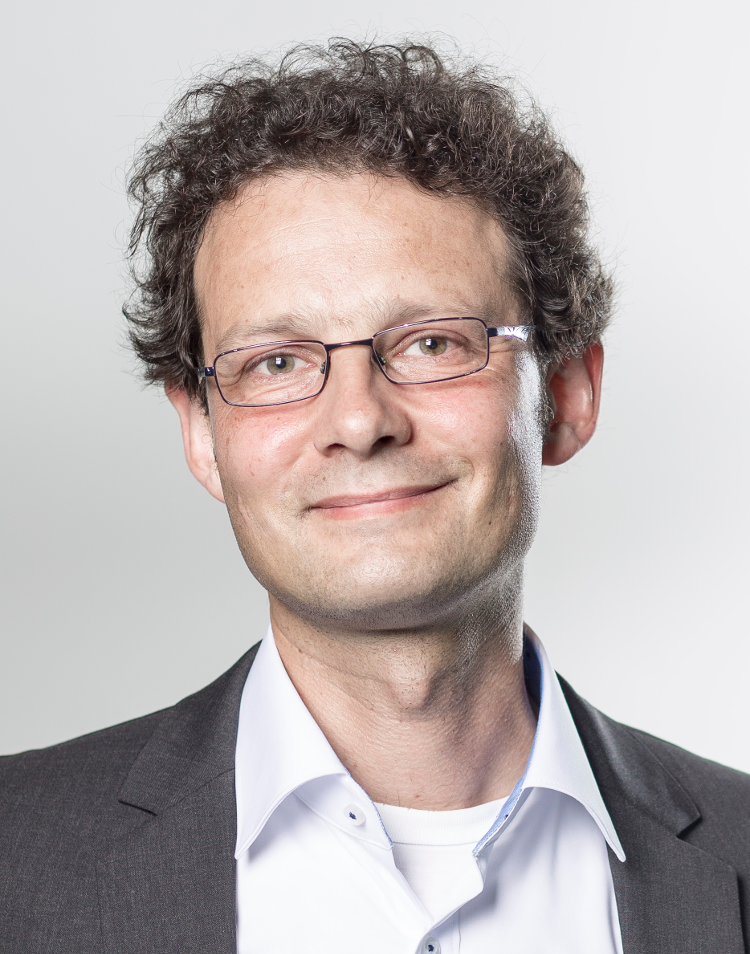}}
\noindent {\bf Dr. Andreas W. Kempa-Liehr} is a Senior Lecturer at the Department of Engineering Science of the University of Auckland, New Zealand, and an Associate Member of the Freiburg Materials Research Center (FMF) at the University of Freiburg, Germany.
Andreas received his doctorate from the University of M\"{u}nster in 2004 and continued his research as head of service group Scientific Information Processing at FMF. 
From 2009 to 2016 he was working in different data science roles at EnBW Energie Baden-W\"{u}rttemberg AG and as Senior Data Scientist at Blue Yonder GmbH.
His research is focussed on the synergy of data science and operations research for engineering applications.
\vspace{1cm}

\parpic{\includegraphics[width=1in,clip,keepaspectratio]{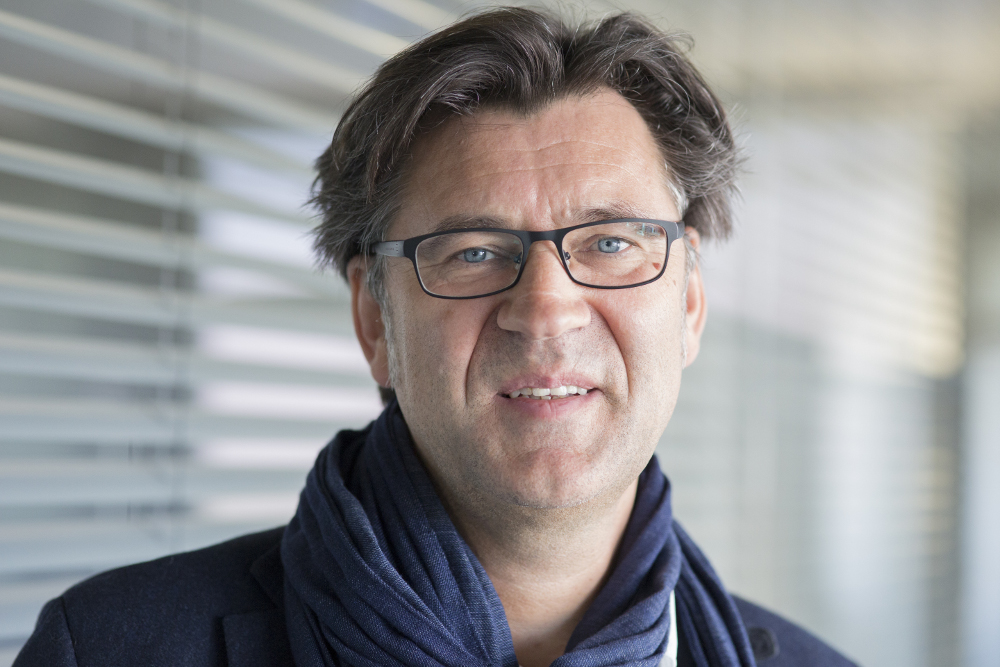}}
\noindent {\bf Prof. Dr. Michael Feindt} founded Blue Yonder GmbH, one of the world’s leading companies for Predictive Applications in the retail sector, in 2008 and serves as its Chief Scientific Advisor and Member of Advisory Board. 
He was a research fellow and staff member at the European Organization for Nuclear Research CERN from 1991 to 1997 and he also participated in the CDFII experiment at Fermilab in Chicago/USA as well as Belle and Belle II experiments at KEK in Japan.  
He is a professor at the Institute of Experimental Nuclear Physics at the Karlsruhe Institute of Technology (KIT), Germany.
His main research interest lies in optimizing and automating research in many different areas
by means of machine learning, especially combining Bayesian statistics, neural networks,
conditional probability density estimation and causality.

\newpage

\appendix
\section{Considered feature mappings}
\label{list_of_feature_mappings}
\alg{FRESH} and the respective \alg{tsfresh} package depend on feature mappings $\theta_k:\mathbb{R}^{n_t}\rightarrow\mathbb{R}$ for capturing relevant characteristics from the respective time series.
Each of these features mappings $\theta(S)$ take a single time series
\begin{equation*}\label{ts_simple}
\begin{split}
S&=(s(t_1), s(t_2), \ldots, s(t_\nu),\ldots,s(t_{n_t}))^{\text{T}}\\
&=  (s_1, \ldots, s_\nu,\ldots,s_{n_t})^{\text{T}}.
\end{split}
\end{equation*}
as argument and return either a real valued feature (e.g. \texttt{length}), or a vector of features (e.g. \texttt{binned\_entropy}).
In the latter case, each element of the returned feature vector will be treated as separate feature.
For the purpose of simplicity, the following descriptions are omitting the indices for different devices and sensors described in Eq.~\eqref{ts_sampling}.
Feature mappings with additional parameters are denoted by $\theta(S|\cdot)$.

Note, that the number of feature mappings provided by the \alg{tsfresh} package is continuously increasing.
However, the following subsections describe the feature mappings, which have been used for the evaluation in Section~\ref{Sec:evaluation}. The most recent list of feature mappings can be found at \url{http://tsfresh.readthedocs.io/en/latest/}.

\subsection{Features from summary statistics}

\begin{itemize}
\item \ftr{maximum}:
  Sample maximum of time series $S$
  \begin{equation*}
    \mathftr{maximum} = \max\; \{s_1,\ldots,s_\nu,\ldots,s_{n_t}\}.
   \end{equation*}

\item \ftr{minimum}:
Sample minimum of time series $S$
  \begin{equation*}
    \mathftr{minimum} = \min\; \{s_1,\ldots,s_\nu,\ldots,s_{n_t}\}.
   \end{equation*}

\item \ftr{mean}:
Arithmetic mean of time series $S$ 
  \begin{equation*}
    \mathftr{mean} = \bar{S} = \frac{1}{n_t}\sum\limits_{\nu=1}^{n_t}s_\nu.
   \end{equation*}

\item \ftr{var}: Expectation of the squared deviation of time series $S$ from its mean $\bar{S}$ without bias correction
  \begin{equation*}
    \mathftr{var} = \frac{1}{n_t}  \sum_{\nu=1}^{n_t}(s_{\nu}-\bar{S})^{2}.
    \end{equation*}

\item \ftr{std}:
  Uncorrected sample standard deviation of time series $S$
  \begin{equation*}
    \mathftr{std} = \sqrt{\mathftr{var}}
    \end{equation*}

\item \ftr{skewness}:
  Sample skewness calculated with adjusted Fisher-Pearson standardized moment coefficient \cite{joanes1998comparing}:
  \begin{equation*}
    \mathftr{skewness} =
    \frac{n_{t}^2}{(n_t-1)(n_t-2)} \frac{\frac{1}{n_t} \sum_{\nu=1}^{n_t} \left(s_\nu - \overline S \right)^3}{\left ( \frac{1}{n_t-1} \sum_{\nu=1}^{n_t} \left(s_\nu - \overline S \right)^2 \right )^{\frac{3}{2}}}.
      \end{equation*}

\item \ftr{kurtosis}: Fourth central moment of time series $S$ divided by the square of its variance
  \begin{equation*}
    \mathftr{kurtosis} =
    \frac{1}{n_t} \sum_{\nu=1}^{n_t} \left(\frac{s_\nu-\bar{S}}{\mathftr{std}}\right)^4 -3
   \end{equation*}
as defined by Fisher \cite{weisstein2002crc}. 
The subtrahend of 3 ensures that the kurtosis is zero for normal distributed samples.

\item \ftr{length}:
  Number of samples $n_t$ of time series $S$
    \begin{equation*}
    \mathftr{length} = n_t.
    \end{equation*}

\item \ftr{median}:
For a time series $S$ with an uneven number of samples $n_t$, the median is the middle of the sorted time series values. 
If the time series has an even number of samples, the two middle values are averaged
\begin{equation*}
  \mathftr{median} =
  \begin{cases} 
      s_{(n_t+1)/2} & : n_t\;\text{is uneven}, \\
      \frac{1}{2}(s_{(n_t/2)}+s_{(n_t/2+1)}) & : n_t\;\text{is even}. \\
   \end{cases}
  \end{equation*}

\item \ftrpar{quantile\_of\_empiric\_distribution\_function}{q}: 
  $q$-quantile $Q_q(S)$ of empirical distribution function $\hat S_{n_t}$ of time series sample $S$
  \begin{equation*}
    Q_q(S)=\text{inf}\left \{  z \ \bigg |  \ \hat S_{n_t} (z ) = \frac{1}{n_t} \sum_{\nu=1}^{n_t} \mathbf{1}_{s_\nu \le z} \ \ge q \ \right \}.
    \end{equation*}
Here, q\% of the ordered values from $S$ are lower or equal to this quantile.
\end{itemize}

\subsection{Additional characteristics of sample distribution}
\begin{itemize}
\item \ftr{absolute\_energy}: Interpreting the time series $S$ as the velocity of a particle with unit mass 2, the observed energy computes to
  \begin{equation*}
    \mathftr{absolute\_energy} = \sum_{\nu=1}^{n_t} s_\nu^2.
    \end{equation*}

\item \ftr{augmented\_dickey\_fuller\_test\_statistic}:
The Augmented Dickey-Fuller test checks the hypothesis that a unit root is present in a time series sample \cite{fuller2009introduction}.
This feature calculator returns the value of the respective test statistic calculated on $S$ \cite{mackinnon1994approximate}. 

\item \ftrpar{binned\_entropy}{m}: 
This feature calculator bins the values of the time series sample into $m$ equidistant bins.
The returned features is 
$$ \sum_{k=0}^{\min(m, n_t)} p_k log(p_k) \cdot \mathbf{1}_{(p_k > 0)}$$ 
where $p_k$ is the percentage of samples in bin $k$.

\item \ftr{has\_large\_standard\_deviation}:
  Boolean feature indicating that
  the standard deviation $\text{std}(S)$ is bigger than half the difference between the maximal and minimal value 
$$\mathftr{std} > \frac{\mathftr{maximum} - \mathftr{minimum}}{2}.$$

\item \ftr{has\_variance\_larger\_than\_std}:  
Boolean feature indicating that the variance is greater than the standard deviation
$$ \mathftr{var} > \mathftr{std},$$
which is equal to the variance of the sample being larger than 1.

\item \ftr{is\_symmetric\_looking}:
Boolean feature indicating 
$$ \left | \mathftr{mean} - \mathftr{median} \right |< \frac{\mathftr{maximum} - \mathftr{minimum}}{2}.$$

\item \ftrpar{mass\_quantile}{q}: 
Relative index $\nu$ where $q \cdot 100\%$ of the mass of the time series $S$ lie left of $\nu$:
$$ \frac{i}{n_t} \quad \text{ such that } \quad i = \min \left \{ k \ \bigg | 1 \le k \le n_t, \  \frac{ \sum_{\nu=1}^{k} s_\nu}{ \overline{S} } \ge q \right \}  .$$
For example for $q = 50\%$ this feature calculator will return the mass center of the time series.

\item \ftr{number\_data\_points\_above\_mean}:
Number of data points, which are larger than the average value of the time series sample. 

\item \ftr{number\_data\_points\_above\_median}:
Number of data points, which are larger than the median value of the time series sample. 

\item \ftr{number\_data\_points\_below\_mean}:
Number of data points, which are lower than the average value of the time series sample. 

\item \ftr{number\_data\_points\_below\_median}:
Number of data points, which are lower than the median value of the time series sample. 

\end{itemize}

\subsection{Features derived from observed dynamics}

\begin{itemize}

\item \ftrpar{arima\_model\_coefficients}{i,k}: 
This feature calculator fits the unconditional maximum likelihood of an autoregressive $AR(k)$ process \cite{jones1980maximum}
$$ s_{\nu}=\varphi_0 +\sum _{{j=1}}^{k}\varphi_{j}s_{{\nu-j}}+\varepsilon_{t}$$
 on the time series $S$.
The parameter $k$ is the maximum lag of the process and the calculated feature is the coefficient $\varphi_{i}$ for index $i$ of the fitted model.

\item \ftrpar{continuous\_wavelet\_transformation\_coefficients}{a,b}: 
Calculates a discretization of the continuous wavelet transformation \cite{du2006improved} for the Ricker wavelet, also known as the \emph{Mexican hat wavelet} \cite{ricker1953form} which is defined by 
$$ \psi(\nu, a, b) = \frac{2}{\sqrt{3a} \pi^{\frac{1}{4}}} \left (1 - \frac{(\nu-b)^2}{a^2} \right) \exp \left (-\frac{(\nu-b)^2}{2a^2} \right).$$
Here, $a$ is the width and $b$ is the location parameter.
The continuous wavelet transformation performs a convolution of the time series $S$ with the wavelet $ \psi(\nu, a, b) $
$$ X_w(a,b)=\sum _{{\nu=-\infty }}^{{\infty }}s_\nu \ \overline \psi \left(\nu, a, b\right).$$
The features are the coefficients $X(a,b)$ for the wavelet of length $a$ at position $b$.

\item \ftrpar{fast\_fourier\_transformation\_coefficient}{k}:
Calculates the one-dimensional discrete Fourier Transform for real input.
The real part of the coefficients
$$ X_{k}\ =\ \sum _{\nu=1}^{n_t}s_{\nu}\cdot \exp \left( -\frac{2\pi i k (\nu-1)}{n_t} \right),\quad k\in \mathbb {Z}$$
are returned as features.
For real valued input the real part of $X_{k}$ is equal to the real part of $X_{-k}$ which means that this feature calculator only take natural numbers as parameters.

\item \ftr{first\_index\_max}:
This feature is the relative position
$$\frac{ \min \left( \arg\max S \right) }{n_t}$$
for which the maximum value was observed for the first time in the time series sample.

\item \ftr{first\_index\_min}:
This feature is the relative position
$$\frac{ \min \left( \arg\min S \right) }{n_t}$$
for which the minimal value was observed for the first time in the time series sample.

\item \ftrpar{lagged\_autocorrelation}{l}: 
Calculates the autocorrelation of the time series $S$ with its lagged version of lag $l$:
$$ \frac{1}{\mathftr{var}} \sum_{\nu=1}^{n_t-l}\left (s_{\nu}-\bar{S}\right)\left(s_{\nu+l}-\bar{S}\right). $$

\item \ftrpar{large\_number\_of\_peaks}{l,m}:
Boolean variable indicating wether the number of peaks of size $l$ as defined in \alg{number\_peaks\_of\_size($l$)} is greater than $m$.

\item \ftr{last\_index\_max}: 
Relative position
$$\frac{ \max \left( \arg\max S \right) }{n_t}$$
at which the maximum value was observed for the last time in the time series sample.

\item \ftr{last\_index\_min}:
  Relative position
$$\frac{ \max \left( \arg\min S \right) }{n_t}$$
  at which the maximum value was observed for the last time in the time series sample.

\item \ftr{longest\_strike\_above\_mean}:
Length of the longest consecutive subsequence in $S$, which is larger or equal to the mean of $S$.

\item \ftr{longest\_strike\_above\_median}: 
Length of the longest consecutive subsequence in $S$, which is larger or equal to the median of $S$.

\item \ftr{longest\_strike\_below\_mean}: 
Length of the longest consecutive subsequence in $S$, which is smaller or equal to the mean of $S$.

\item \ftr{longest\_strike\_below\_median}: 
Length of the longest consecutive subsequence in $S$, which is smaller or equal to the median of $S$.

\item \ftr{longest\_strike\_negative}: 
Length of the longest consecutive subsequence in $S$, which is smaller or equal to the median of $S$.

\item \ftr{longest\_strike\_positive}: 
Length of the longest consecutive subsequence in $S$ that is smaller or equal to the median of $S$.

\item \ftr{longest\_strike\_zero}: 
Length of the longest consecutive subsequence in $S$ that is smaller or equal to the median of $S$.

\item \ftr{mean\_absolute\_change}:
Arithmetic mean of absolute differences between subsequent time series values 
$$ \frac{1}{n_t} \sum_{\nu=1}^{n_t-1} | s_{\nu+1} - s_{\nu}|. $$

\item \ftrpar{mean\_absolute\_change\_quantiles}{q_l,q_h}: 
This feature calculator first fixes a corridor given by the quantiles $Q_{q_l}$ and $Q_{q_h}$ of the empirical distribution function of $S$.
Then it calculates the average absolute value of consecutive changes of the series $S$ inside this corridor: 
$$ \frac{\sum_{\nu=1}^{n_t-1} | s_{\nu+1} - s_{\nu}| \ \textbf{1}_{Q_{q_l} \le s_\nu \le Q_{q_h}} \ \textbf{1}_{Q_{q_l} \le s_{\nu+1} \le Q_{q_h}}}{\sum_{\nu=1}^{n_t-1} \textbf{1}_{Q_{q_l} \le s_\nu \le Q_{q_h}} \textbf{1}_{Q_{q_l} \le s_{\nu+1} \le Q_{q_h} }}.$$

\item \ftr{mean\_autocorrelation}:
Average autocorrelation over possible lags $l$ ranging from $1$ to $n_t-1$:
$$ \frac{1}{(n_t-1)\var(S)} \sum_{l=1}^{n_t} \sum_{\nu=1}^{n_t-l}(s_{\nu}-\overline{S})(s_{\nu+l}-\overline{S}).$$

\item \ftr{mean\_second\_derivate\_central}:
Average value of the second derivate of the time series  
$$ \frac{1}{n_t-2} \sum_{\nu=2}^{n_t-2} \frac{1}{2} (s_{\nu-1} - 2 \cdot s_\nu + s_{\nu+1}) .$$

\item \ftrpar{number\_continous\_wavelet\_transformation\_peaks\_of\_size}{l}: 
This feature calculator detects peaks in the time series by inspecting the coefficients of a discretization of the continuous wavelet transformation of the time series $S$.
First, it calculates the wavelet transformation $X_w(a,b)$ for widths $a$ ranging from 1 to $l$.
Again the ricker wavelet is used.
The peaks are then found by comparing the local signal-to-noise ratio (SNR), if a local maxima of the wavelet coefficients $X_w(a,b)$ is observed for enough widths, it is counted as a peak \cite{du2006improved}.

\item \ftrpar{number\_peaks\_of\_size}{l}:
A peak of support $l$ is defined as a data point $s_i$, which is larger than its $l$ neighboring values to the left and to the right.
This feature calculates the number of peaks with support of $l$ in the time series $S$.

\item \ftrpar{spektral\_welch\_density}{i}:
This feature calculator uses Welch's method \cite{welch1967use} to compute an estimate of the power spectral density of the time series sample $S$.
Welch's methods divides the time series into overlapping segments, computes a modified periodogram for each segment and then averages the periodograms.
The frequencies $f$ are defined as follow 
\begin{equation*}
\begin{split}
f &= \left ( 0, \frac{1}{n_t}, \frac{2}{n_t}, \ldots, \frac{1}{2}-\frac{1}{n_t}, -\frac{1}{2}, - \frac{1}{2}+\frac{1}{n_t}, \ldots,-\frac{2}{n_t}, -\frac{1}{n_t}\right) \quad \text{if $n_t$ is even} \\
f &= \left ( 0, \frac{1}{n_t}, \frac{2}{n_t}, \ldots, \frac{n_t -1}{2n_t}, -\frac{n_t -1}{2n_t}, -\frac{1}{2}+\frac{1}{n_t}, \ldots,-\frac{2}{n_t}, -\frac{1}{n_t}\right) \quad \text{if $n_t$ is odd}
\end{split}
\end{equation*}
This feature calculator will return the calculated power spectrum of frequency $f_i$ as the feature. 

\item \ftrpar{time\_reversal\_asymmetry\_statistic}{l}:
Value of
$$ \frac{1}{n_t-2l} \sum_{\nu=1}^{n_t-2l} s_{\nu + 2 l}^2 \cdot s_{\nu + l} - s_{\nu + l} \cdot  s_{\nu}^2$$
for lag $l$.
It was proposed in \cite{fulcher_highly_2014} as promising feature to extract from time series.

\end{itemize}

\end{document}